\documentclass[10pt,twocolumn,letterpaper]{article}

\usepackage[pagenumbers]{wacv}   % de-anonymized, with page numbers, for arXiv    % To produce the REVIEW version for the algorithms track
\usepackage{algorithm}
\usepackage{algpseudocode}
\usepackage{multirow}

\definecolor{wacvblue}{rgb}{0.21,0.49,0.74}
\usepackage[pagebackref,breaklinks,colorlinks,allcolors=wacvblue]{hyperref}
\usepackage{hyperref} 

\def\wacvPaperID{1358} % *** Enter the WACV Paper ID here
\def\confName{WACV}
\def\confYear{2027}
\usepackage{xspace}
\newcommand{\method}{LAD\xspace}
\newcommand{\methodfull}{Language-Anchored Decomposition\xspace}
\usepackage{booktabs}
\usepackage{siunitx}
\usepackage{booktabs}
\usepackage{siunitx}
\title{Naming the Concepts Classifiers Rely On: Language-Anchored Decomposition for Faithful Explanation }

\author{%
Ahsan Habib Akash$^{1}$ \quad
Dipkamal Bhusal$^{2}$ \quad
Stacey Jones$^{3}$ \\
Donald A. Adjeroh$^{1}$ \quad
Binod Bhattarai$^{4,5,6}$ \quad
Prashnna Kumar Gyawali$^{1}$\thanks{Corresponding author: \texttt{prashnna.gyawali@mail.wvu.edu}} \\[4pt]
$^{1}$West Virginia University, USA \quad
$^{2}$Rochester Institute of Technology, USA \quad
$^{3}$O Analytics \\
$^{4}$University of Aberdeen, UK
 \quad
  $^{5}$Fogsphere (Redev.AI Ltd, UK) \quad  \\
  $^{6}$ University College London, UK
}

\begin{document}
\maketitle
\begin{abstract}
Deep neural networks are widely deployed in high-stakes visual applications
where interpretability is critical, yet existing explanations face a trade-off:
post-hoc concept methods recover factors that are faithful to a model's behavior
but unnamed, while naming and by-design methods attach human-readable concepts
only by retraining or altering the classifier. We propose \methodfull{}
(\method{}), a post-hoc framework that delivers concepts which are simultaneously
named, faithful, and obtained without modifying the model. For each class, a
large language model proposes a concept vocabulary that CLIP-based similarity
maps localize across image regions. Inverting standard non-negative matrix
factorization, \method{} fixes these language-grounded maps as the coefficient
matrix and learns only a concept basis that reconstructs the frozen encoder's
activations, so naming becomes a structural constraint and the model's own
feature geometry determines which concepts are retained. Removing this anchor
preserves accuracy but collapses attribution faithfulness. Across natural-image, scene, and medical-imaging benchmarks,
\method{} produces spatially precise explanations that are decision-relevant
under both concept insertion and deletion, while uniquely providing
stable, human-interpretable concept names.
Code is available at: 
\url{
https://github.com/machine-intelligence-lab-wvu/LAD}.
\end{abstract}    
\section{Introduction}
\label{sec:intro}

\begin{figure}[t]
    \centering
    \includegraphics[width=\linewidth]{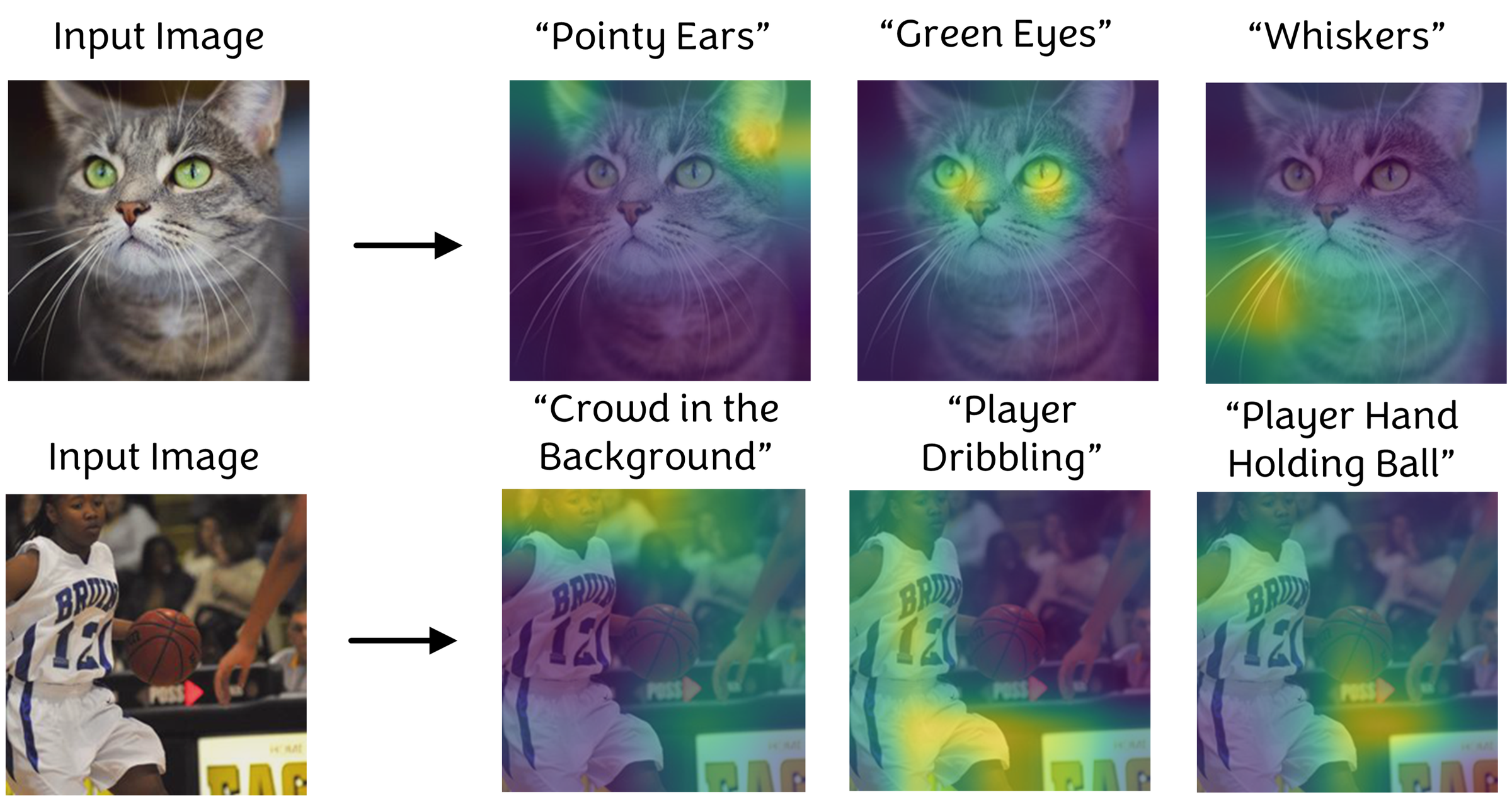}
   \caption{Qualitative illustration of \method{}. Each image is decomposed into
spatially localized, named concepts, including part-level object concepts (e.g.,
``Pointy Ears,'' ``Green Eyes'') and contextual scene concepts
(e.g., ``Crowd in the Background,'' ``Player Dribbling''). Because each concept is
anchored to language before decomposition, the explanations are both
human-interpretable and faithful to the frozen classifier's predictions.}
    \label{fig:concept_decomposition}
\end{figure}

\noindent Interpreting the decision mechanisms of deep visual models is crucial for
diagnosing failures, mitigating biases, and ensuring responsible deployment in
high-stakes applications. Post-hoc attribution methods such as
Grad-CAM~\cite{selvaraju2017grad} and Integrated
Gradients~\cite{sundararajan2017axiomatic} visualize the input regions most
influential to a prediction, but they operate at the pixel level and emphasize
low-level activations rather than the high-level cues humans reason with, offering
limited semantic insight into a model's decision
process~\cite{nguyen2021effectiveness}. Concept-based explanations address this
by attributing predictions to human-interpretable units such as object parts
and visual attributes~\cite{kim2018interpretability,fel2023craft,ghorbani2019towards}.

\noindent Existing concept-based methods, however, fall into two families that each pay a
price. The first is \emph{post-hoc concept discovery}. Building on
TCAV~\cite{kim2018interpretability}, unsupervised methods such as
ACE~\cite{ghorbani2019towards}, ICE~\cite{zhang2021ice}, and
CRAFT~\cite{fel2023craft} extract concepts via clustering or Non-Negative
Matrix Factorization (NMF), and FACE~\cite{bhusal2025face} further aligns the
decomposition with classifier logits to improve faithfulness. These methods are
faithful to the model, but the recovered factors are unnamed: the same factor
index can correspond to different visual elements across images, so a human must
label each factor after the fact and consistency is not guaranteed. The second
family attaches names directly. Concept Bottleneck
Models~\cite{koh2020concept}, their label-free
variants~\cite{oikarinen2023labelfree}, and spatial extensions such as Show and
Tell~\cite{benou2025show} predict named concepts before classification, while
neuron-naming methods such as LaViSE~\cite{yang2022lavise} train an auxiliary
explainer to describe individual filters. These methods
produce names, but at a cost: by-design models replace the classifier head with
a newly trained predictor, so the explained model is no longer the deployed
one~\cite{rudin2019stop}, and neuron-naming methods describe units without
testing whether the named concepts are actually responsible for predictions. No
prior method delivers concepts that are simultaneously \emph{named},
\emph{faithful}, and obtained \emph{without modifying the model}.

\noindent We fill this gap with \methodfull{} (\method{}) by inverting the standard
decomposition. For each target class, a large language model proposes a
class-specific concept vocabulary; for \textit{``cat''}, the proposals may
include ``pointy ears,'' ``green eyes,'' and ``whiskers.'' We then use
CLIP~\cite{radford2021clip} image--text similarity to localize each concept
across image regions. Rather than learning latent coefficients as in NMF,
\method{} \emph{fixes} these language-grounded similarity maps as the coefficient
matrix and learns only a concept basis that reconstructs the frozen encoder's
activations. Naming therefore becomes a structural constraint rather than a
post-hoc label, and the model's own feature geometry, not CLIP alone, determines
which named concepts are retained. \method{} needs no auxiliary training and no
concept annotations, and tests faithfulness by reconstruction rather than
describing units in isolation.

\noindent Crucially, the language anchor is not cosmetic. Replacing the fixed
language-grounded coefficients with learned ones, which recovers standard
unsupervised NMF on the same activations, preserves classification accuracy but
sharply reduces attribution faithfulness.
(\cref{sec:ablation}). The anchor changes \emph{which} basis directions are
discovered, steering them toward classifier-meaningful structure, and the
reconstruction objective reshapes the initial CLIP similarities so that concepts
inconsistent with the encoder's evidence are suppressed, mitigating known biases
in CLIP-style models. \cref{fig:concept_decomposition} illustrates the resulting
explanations, which recover fine-grained part-level concepts (top) and
higher-level scene structure (bottom), each localized and paired with a
natural-language name.

\noindent \underline{Main Contributions:}
\begin{itemize}
    \item A post-hoc concept discovery framework, \methodfull{} (\method{}),
    that inverts NMF by fixing language-grounded CLIP similarity maps as the
    coefficient matrix and learning only the concept basis against frozen
    encoder activations. Naming is enforced structurally, with no retraining and
    no manual concept annotations.
    \item Evidence that the language anchor is functional rather than cosmetic:
    removing it preserves accuracy but collapses deletion faithfulness, showing
    that language supervision determines which concept directions are discovered.
    
    \item Evaluation across natural-image, scene, and medical benchmarks
    showing that \method{}'s named concepts are decision-relevant under concept
    insertion and deletion, at faithfulness comparable to the strongest prior
    method.
\end{itemize}
\section{Related Work}
\label{sec:related}

Interpretability methods fall into two broad approaches: \textit{post-hoc}
methods that interpret a model after training, and \textit{ante-hoc}
explanation-by-design architectures that constrain the model during learning.

\noindent
\textbf{Feature attribution:}
Saliency maps~\cite{simonyan2014deep}, Grad-CAM~\cite{selvaraju2017grad},
Integrated Gradients~\cite{sundararajan2017axiomatic}, and perturbation-based
variants~\cite{smilkov2017smoothgrad,petsiuk2018rise} attribute importance to
input pixels. These pixel-level maps lack semantic interpretability and can be
visually plausible yet unfaithful to the model's
decision~\cite{colin2022cannot,nguyen2021effectiveness}.

\noindent
\textbf{Post-hoc concept discovery:}
TCAV~\cite{kim2018interpretability} quantifies sensitivity to user-defined
concepts but needs manual annotation, while ACE~\cite{ghorbani2019towards},
ICE~\cite{zhang2021ice}, and CRAFT~\cite{fel2023craft} automate discovery via
clustering or Non-negative Matrix Factorization (NMF).
FACE~\cite{bhusal2025face} improves faithfulness with a KL regularizer aligning
reconstructed and original predictions. These methods are faithful but their
factors are \emph{unnamed}: a factor index may map to different visual elements
across images, so each must be labeled after the fact with no consistency
guarantee.

\noindent
\textbf{Naming individual units:}
A parallel line attaches language to model internals at the level of individual
neurons or filters: Network Dissection~\cite{bau2017network} via overlap with
annotated concept masks, and MILAN~\cite{hernandez2022milan} and
CLIP-Dissect~\cite{oikarinen2023clipdissect} via open-vocabulary descriptions of
top-activating regions. LaViSE~\cite{yang2022lavise} maps masked filter
activations to a semantic space to describe filters, including unseen concepts.
These methods differ from ours on two axes. First, granularity: they name single
units, whereas decomposition-based discovery, including \method{}, recovers
concepts as \emph{directions} in activation space (combinations of units), the
unit at which NMF-style methods operate. Second, faithfulness: they produce
\emph{descriptive} labels without testing whether a named concept is causally
responsible for a prediction, and several~\cite{bau2017network,hernandez2022milan,yang2022lavise}
further require dense annotations or a trained explainer.

\noindent
\textbf{Concept bottleneck and by-design models:}
Concept Bottleneck Models and their label-free and spatial
variants~\cite{koh2020concept,losch2019interpretability,oikarinen2023labelfree,wang2023learning,benou2025show}
predict named concepts before the output, recently using vision--language models
to avoid manual labels. They are interpretable by construction but replace the
classifier head with a newly trained predictor, so the explained model is no
longer the deployed one~\cite{rudin2019stop}.

\noindent
\textbf{Our position:}
Post-hoc discovery is faithful but unnamed, unit-naming provides names without
verifying predictive responsibility, and by-design models attach names only by
retraining. \method{} occupies the intersection these families leave open: named,
faithful, and obtained without modifying the model, by fixing language-grounded
coefficients and learning the basis against the frozen encoder.

\label{sec:method}
\begin{figure*}[t]
    \centering
    \includegraphics[width=.85\textwidth]{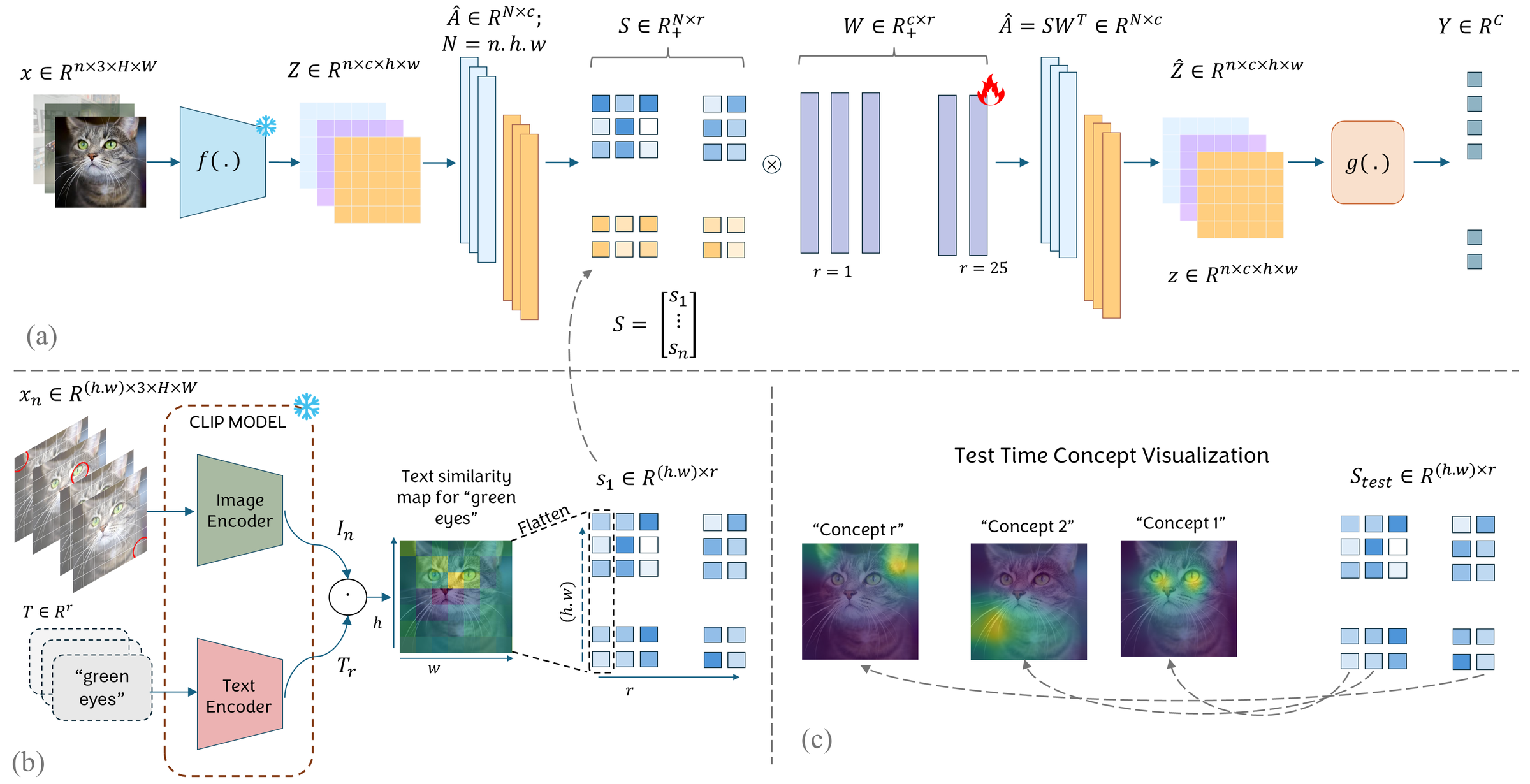}
    \caption{
        \textbf{\method{} framework.}
        (a) An input image is processed through a pretrained encoder $f(\cdot)$
        to obtain latent activation maps, divided into spatial patches and
        arranged into a matrix $\bar{A} \in \mathbb{R}^{N \times p}$.
        (b) Using CLIP-based semantic priors, the concept similarity matrix $S$
        is \emph{fixed} as the coefficient matrix while the basis $W$ is learned
        through a language-anchored reconstruction $\bar{A} \approx S W^{\top}$;
        the reconstructed activations preserve spatial coherence and predictive
        alignment with the classifier head $g(\cdot)$.
        (c) At inference, the estimated concept activations yield interpretable
        spatial maps and class-specific concept importance vectors.
    }
    \label{fig:method}
\end{figure*}
\section{Methodology}

An overview of \method{} is shown in Fig.~\ref{fig:method}. We describe it
beginning with the problem setup.

\subsection{Setup}
\label{sec:Setup}

A pretrained model maps inputs from an image space
$\mathcal{X}\subset\mathbb{R}^{3\times H\times W}$ to predictions in
$\mathcal{Y}\subset\mathbb{R}^{c}$. We decompose it into an \emph{encoder}
$f:\mathcal{X}\rightarrow\mathbb{R}^{p\times h\times w}$ that produces spatial
feature maps and a \emph{classifier head} $g:\mathbb{R}^{p}\rightarrow\mathbb{R}^{c}$
on globally pooled features. Both remain \emph{frozen}: \method{} explains the
deployed model without retraining. For a batch of $n$ inputs $\mathbf{X}$, the
encoder produces
\[
\mathbf{Z}=f(\mathbf{X})\in\mathbb{R}^{n\times p\times h\times w},
\]
where $p$ is the channel dimension and $(h,w)$ the spatial resolution. Global
average pooling (GAP) over the spatial dimensions yields sample-wise latents, with
predictions $g(\mathbf{A})$:
\[
\mathbf{A}=\operatorname{GAP}(\mathbf{Z})\in\mathbb{R}^{n\times p}.
\]

\vspace{4pt}
\noindent
\textbf{Post-hoc Concept Discovery:}
Given a frozen model $g \circ f$, post-hoc concept discovery seeks a
low-dimensional, non-negative factorization of the latents $\mathbf{A}$ into a
small set of interpretable components. Most methods operate on pooled activations
$\mathbf{A}$, restricting analysis to a single class to avoid cross-class
confounds. Non-negative Matrix Factorization (NMF) is the common formulation:
\[
\mathbf{A} \;\approx\; \mathbf{U}\mathbf{W}^{\top},
\qquad
\mathbf{U}\in\mathbb{R}^{n\times r}_{+},\;\;
\mathbf{W}\in\mathbb{R}^{p\times r}_{+},
\]
where $r$ is the number of concepts. Each column of $\mathbf{W}$ defines a concept
direction in feature space, and the matching column of $\mathbf{U}$ its
per-sample activation strength. Reconstructed features
$\widehat{\mathbf{A}}=\mathbf{U}\mathbf{W}^{\top}$ can be passed through $g(\cdot)$
to assess predictive consistency.

\subsection{Semantically Anchored Matrix Factorization}
\label{sec:semantically_aligned_factorization}

Unlike methods that factorize globally pooled representations
$\mathbf{A}\in\mathbb{R}^{n\times p}$, we factorize directly in the encoder's
spatial feature space to preserve localized evidence. We unfold the activations
$\mathbf{Z}$ over spatial and batch dimensions:
\[
\bar{\mathbf{A}} = \operatorname{Unfold}(\mathbf{Z}) \in \mathbb{R}^{(n h w)\times p},
\]
so each row is a spatial location and each column a feature channel. Pooling
before factorization implicitly assumes concepts are spatially homogeneous across
an image; unfolding instead treats each location as an independent observation,
raising the effective sample size from $n$ to $n h w$ and capturing localized,
part-level structure that pooling would average out. Concept discovery thus
becomes a structured clustering over spatial feature vectors while remaining
consistent with the classifier's global decision pipeline.

\noindent
\textbf{Inverting the Coefficient Matrix:}
Standard NMF, $\mathbf{A} \approx \mathbf{U}\mathbf{W}^{\top}$, learns both
$\mathbf{U}$ and $\mathbf{W}$, so the coefficients $\mathbf{U}$ are unconstrained
and lack semantic alignment. We invert this: rather than discovering factors and
labeling them afterward, we \emph{fix} the coefficients to a semantically grounded
matrix $\mathbf{S}$ from language--vision similarity and learn only the basis,
\[
\hat{\bar{\mathbf{A}}} \approx \mathbf{S}\mathbf{W}^{\top},
\qquad
\mathbf{S}\in\mathbb{R}_{+}^{(n h w)\times r}.
\]
Each column of $\mathbf{S}$ is a human-interpretable concept and each entry the
similarity between a spatial location and a concept prompt. This turns
factorization into a semantically constrained reconstruction: we learn a basis
$\mathbf{W}$ that best reconstructs encoder activations under fixed,
language-aligned coefficients. Because the concept identity is fixed a priori by
$\mathbf{S}$, each basis vector is tied to a named concept, and naming becomes a
structural property of the factorization rather than a post-hoc step.

\subsection{Language-Anchored Concept Map Construction}
\label{sec:concept_similarity_construction}

\textbf{Class-Specific Concept Vocabulary:} For each class $c \in \mathcal{C}$ we
build a fixed-size vocabulary $\mathcal{T}_c = \{t_1,\dots,t_r\}$ of visually
grounded concepts ($r=25$ unless noted; sensitivity in \cref{sec:ablation}).
Following~\cite{oikarinen2023labelfree}, we prompt a large language model with
structured templates eliciting diverse attributes (object parts, textures,
shapes, materials, poses, contextual cues). Automatic filtering removes (i)
generic filler terms (e.g., ``object''), (ii) concepts overlapping the class name
(preventing trivial leakage), and (iii) near-duplicates by CLIP text-embedding
similarity, yielding a compact, diverse vocabulary with no manual annotation.
Templates and filtering details are in the supplement.

\noindent
\textbf{Localized CLIP Similarity Maps:} Given an image $x_n$ and its vocabulary
$\mathcal{T}_c$, we construct spatial concept maps via localized
prompting~\cite{shtedritski2023does, benou2025show}. After CLIP's deterministic
preprocessing, we define a grid aligned to $(h,w)$ and, at each location, form a
variant $x_n^{(h,w)}$ by overlaying a small red circle. With $\ell_2$-normalized
CLIP encoders $E_I,E_T$, the cosine similarity for concept $t_m$ is
\[
s_{n,m}^{(h,w)} =
\frac{
    E_I(x_n^{(h,w)}) \cdot E_T(\texttt{`a photo of } t_m \texttt{'})
}{
    \|E_I(x_n^{(h,w)})\|_2 \, \|E_T(\texttt{`a photo of } t_m \texttt{'})\|_2
}.
\]
This gives a spatial map $s_{n,m} \in \mathbb{R}^{h \times w}$; flattening yields
$\mathbf{s}_n \in \mathbb{R}^{(h w)\times r}$ per image, and stacking across images
forms the fixed coefficient matrix
\[
\mathbf{S} = [\,\mathbf{s}_1;\dots;\mathbf{s}_n\,]
\in \mathbb{R}^{(n h w)\times r},
\]
whose rows are spatial locations and columns named concepts, used in the
reconstruction $\hat{\bar{\mathbf{A}}} \approx \mathbf{S}\mathbf{W}^{\top}$.

\subsection{Activation-Level Reconstruction Optimization}

With $\mathbf{S}$ fixed, we recover the concept basis by minimizing the
reconstruction error
\begin{equation}
    \min_{\mathbf{W} \ge 0} \ \mathcal{L}_{\text{recon}} =
    \frac{1}{2}\|\bar{\mathbf{A}} - \mathbf{S}\mathbf{W}^\top\|_F^2.
    \label{eq:recon_loss}
\end{equation}
Since GAP is linear and the head reads only pooled features, a small
reconstruction error in $\bar{\mathbf{A}}$ bounds the perturbation of $\mathbf{A}$,
and hence of the logits; empirically, minimizing Eq.~\ref{eq:recon_loss} alone
preserves accuracy across datasets and backbones without output-level
regularization. Unlike prior approaches that jointly learn both factors or
regularize logits, we fix $\mathbf{S}$ and learn only $\mathbf{W}$, enforcing
semantic alignment structurally rather than through auxiliary losses.

We optimize Eq.~\ref{eq:recon_loss} with projected gradient descent (PGD) to
enforce non-negativity:
\begin{equation}
    \mathbf{W} \leftarrow \max(0, \mathbf{W} - \eta \nabla_{\mathbf{W}}\mathcal{L}_{\text{recon}}),
\end{equation}
initialized with NNDSVD~\cite{boutsidis2008svd}. With $\mathbf{S}$ fixed the
objective is smooth in $\mathbf{W}$ and PGD converges to a stationary point
(supplement).
\begin{figure*}[!t]
    \centering
    \includegraphics[width=\textwidth]{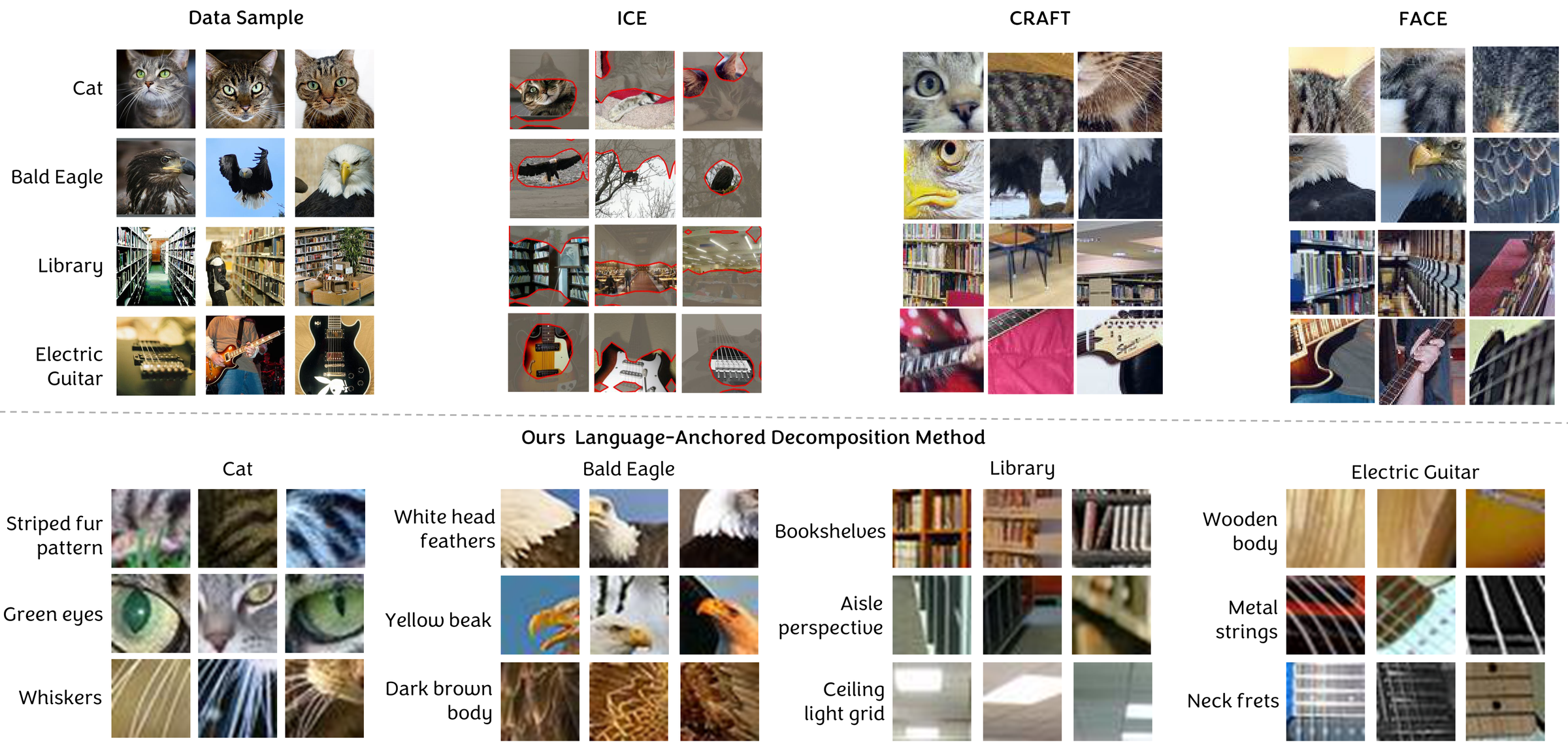}
   \caption{
Concept discovery results on ImageNet. The upper panel shows top-ranked concepts
from NMF baselines (ICE, CRAFT, FACE), which highlight coarse regions without
semantic names; the lower panel shows \method{}, which produces fine-grained,
spatially coherent concepts each paired with a human-interpretable name.
}
    \label{fig:concept_comparison}
\end{figure*}
\subsection{Inference via Non-Negative Semantic Coefficient Estimation}

With the basis $\mathbf{W}$ learned, we explain a new image $x$ against the same
frozen encoder. We extract and unfold its activations to
$\bar{\mathbf{A}} \in \mathbb{R}^{(h w)\times p}$ and, with $\mathbf{W}$ fixed,
estimate non-negative coefficients
$\hat{\mathbf{S}} \in \mathbb{R}^{(h w)\times r}_{+}$ by non-negative least squares:
\begin{equation}
\min_{\hat{\mathbf{S}} \ge 0}
\frac{1}{2}
\left\|
\hat{\mathbf{S}}\mathbf{W}^{\top} - \bar{\mathbf{A}}
\right\|_F^2.
\label{eq:nnls_inference}
\end{equation}
A fast non-negative initialization is the projected normal equation
\begin{equation}
\hat{\mathbf{S}}_0 =
\operatorname{ReLU}\!\left(
\bar{\mathbf{A}}\mathbf{W}(\mathbf{W}^{\top}\mathbf{W})^{-1}
\right),
\end{equation}
optionally refined by projected gradient descent,
\begin{equation}
\hat{\mathbf{S}} \leftarrow
\operatorname{ReLU}\!\left(
\hat{\mathbf{S}} - \eta(\hat{\mathbf{S}}\mathbf{W}^{\top}\mathbf{W} - \bar{\mathbf{A}}\mathbf{W})
\right),
\end{equation}
with $\eta$ from the spectral norm of $\mathbf{W}^{\top}\mathbf{W}$. The
reconstruction $\hat{\bar{\mathbf{A}}}=\hat{\mathbf{S}}\mathbf{W}^{\top}$ is then
reshaped and passed through the head after pooling; predictive preservation is
the accuracy of the original versus reconstructed activations.

\subsection{Concept Heatmaps and Visualization}

We project the inferred coefficients back to the spatial domain. Reshaping
$\hat{\mathbf{S}}$ to $\hat{\mathbf{S}}_{\text{spatial}} \in \mathbb{R}^{h \times w \times r}$,
each concept $k$ gives a normalized heatmap
\begin{equation}
\mathbf{M}_k = \mathrm{Normalize}\!\left(\hat{\mathbf{S}}_{\text{spatial}}(:,:,k)\right),
\end{equation}
rescaled to $[0,1]$, highlighting regions whose features are best reconstructed by
the $k$-th basis vector. Overlaid on the image, and since channel $k$ corresponds
to the named concept $t_k$ (e.g., ``pointy ears''), each map is a semantically
grounded explanation of localized evidence.
% \section{Experiments}

\section{Experimental Setup}
\label{sec:experiments}

\noindent
\textbf{Datasets:}
We evaluate \method{} on ImageNet (ILSVRC 2012) using a publicly available
pretrained ResNet34, analyzing 500 of the 1{,}000 classes, a substantial
expansion over prior concept-discovery work that reports on as few as 10
classes~\cite{bhusal2025face}. For scene-centric generalization we use
Places365-Standard with the publicly available pretrained ResNet50 checkpoint,
covering 364 of 365 categories (one excluded due to $0\%$ baseline accuracy). We
also evaluate a clinical domain (\cref{sec:retina}); additional analyses,
including results on further CNN backbones and a Vision Transformer case study,
are provided in the supplementary material.

\noindent
\textbf{Baselines:}
We compare against representative NMF-based concept discovery methods:
ICE~\cite{zhang2021ice}, CRAFT~\cite{fel2023craft}, and FACE~\cite{bhusal2025face},
all using identical backbones, preprocessing, data splits, and concept counts $r$.

\noindent
\textbf{Metrics:}
\textbf{Acc} is top-1 agreement between predictions on original and reconstructed
activations. \textbf{C-Ins} ($\uparrow$) measures whether concepts are
\emph{sufficient} for the decision (performance restored as top concepts are
inserted), and \textbf{C-Del} ($\uparrow$) whether they are \emph{necessary}
(performance degraded as they are removed).

\section{Results}
\subsection{Named Concepts and Per-Image Rollouts}
\label{sec:rollouts}

\noindent
Because \method{} anchors each factor to a language-defined concept, a concept
refers to the same entity wherever it appears: \emph{maple neck}, \emph{fretted
scale}, and \emph{tuning pegs} localize to the same parts across every image of
the \emph{Electric Guitar} class. NMF baselines (ICE, CRAFT, FACE) recover
\emph{unnamed} factors, so the same factor index may highlight the guitar body in
one image and the fretboard in another, limiting cross-image
comparability~\cite{lipton2018mythos,molnar2020general}. \Cref{fig:concept_comparison} contrasts the two: baseline factors highlight coarse
or discriminative regions with no name, whereas \method{} recovers fine-grained
concepts each tied to a human-interpretable label.

\begin{figure*}[t]
    \centering
    \includegraphics[width=\textwidth]{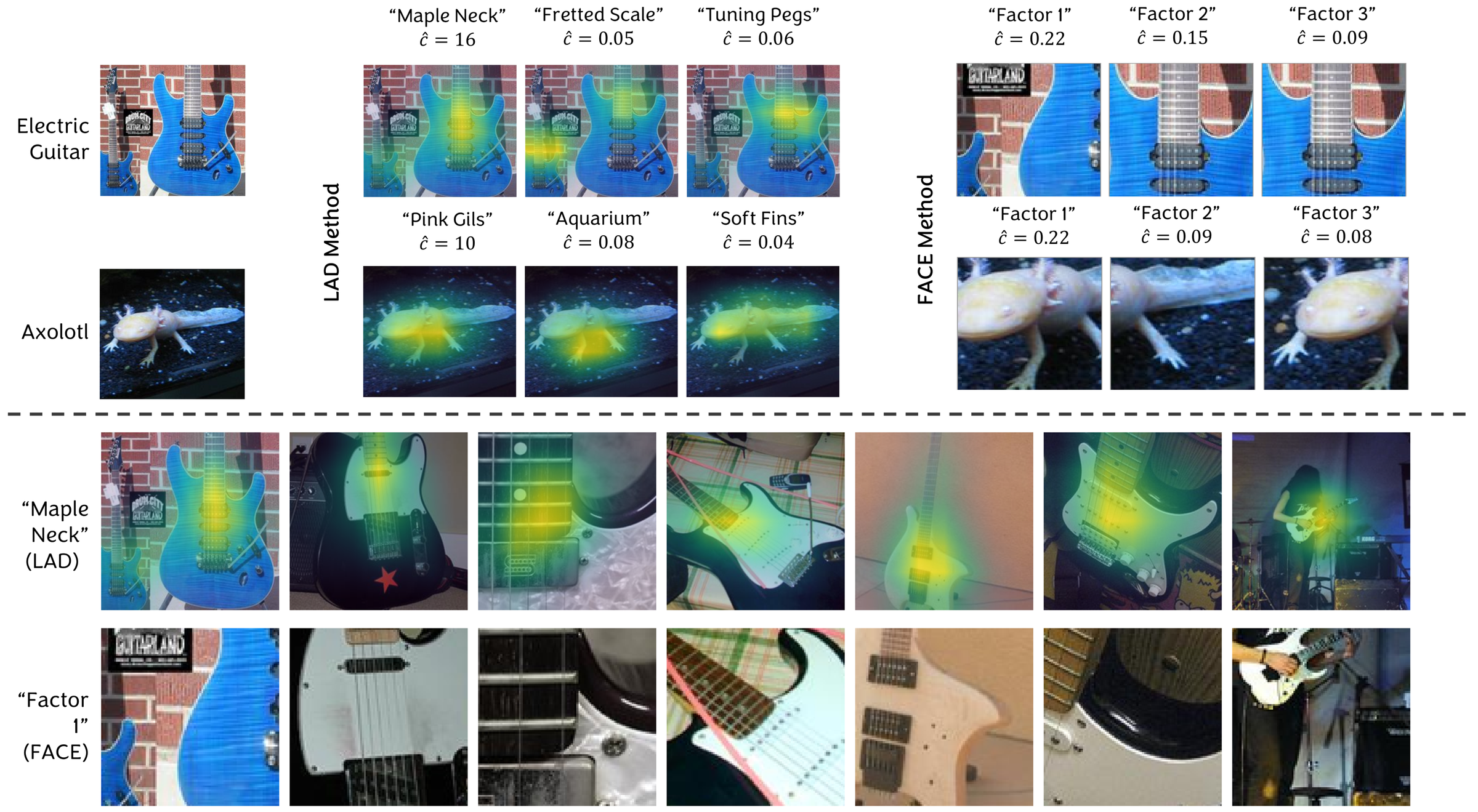}
    \caption{\textbf{Per-image named-concept rollouts, a capability unique to
    \method{}.}
    \textbf{Top:} for an electric guitar ($p{=}0.99$) and an axolotl ($p{=}1.00$),
    \method{} returns localized, \emph{named} concepts with their reconstruction
    contribution $\hat{c}$ (\cref{eq:contribution}); FACE returns factors of
    comparable contribution but with no semantic label (Factor 1--3).
    \textbf{Bottom:} the same named concept (\emph{maple neck}) localizes to the
    neck across different images of the class, whereas FACE's leading factor
    (Factor 1) attaches to a different region in each image, so its meaning is not
    consistent across images. Examples are the median-C-Ins image per class, fixed
    in advance.}
    \label{fig:rollout}
\end{figure*}

\noindent
This naming enables a capability no baseline provides: the \emph{per-image
named-concept rollout}. For a single input, \method{} returns the named concepts
the model used, each localized, ranked by importance, and annotated with its
contribution to the reconstruction,
\begin{equation}
\hat{c}_k \;=\;
\frac{\lVert \hat{\mathbf{S}}[:,k]\rVert_2\,\lVert \mathbf{W}[:,k]\rVert_2}
     {\sum_{j} \lVert \hat{\mathbf{S}}[:,j]\rVert_2\,\lVert \mathbf{W}[:,j]\rVert_2},
\label{eq:contribution}
\end{equation}
the share of the reconstruction explained by concept $k$; the baselines can only
emit ``factor 1, factor 2, \dots'' for a human to label afterward.
\Cref{fig:rollout} (top) shows this for an electric guitar and an axolotl:
\method{} returns localized, named concepts with their contributions, while FACE
returns factors of comparable contribution but no semantic label.

\noindent
Naming also yields \emph{cross-image consistency} that unnamed factors lack.
\Cref{fig:rollout} (bottom) tracks one \method{} concept, \emph{maple neck},
against FACE's leading factor across several images of the class: the named
concept localizes to the neck in every image, whereas the unnamed factor attaches
to a different region each time, so its meaning cannot be read off the index. The
rollout is thus the per-image, named view of the structure our aggregate metrics
summarize (\cref{tab:main_comparison}): the concepts it surfaces are those whose
insertion and deletion move the prediction.

\begin{table}[t]
\centering
\caption{Primary comparison across datasets. All methods preserve accuracy, so
faithfulness is read from concept insertion (C-Ins) and deletion (C-Del):
\method{} leads on insertion and uniquely names its concepts, while FACE is
strongest on deletion. Best per column within each panel in bold.}
\label{tab:main_comparison}
\setlength{\tabcolsep}{12pt}
\begin{tabular}{l *{3}{S[table-format=1.3]}}
\toprule
Method & {Acc $\uparrow$} & {C-Ins $\uparrow$} & {C-Del $\uparrow$} \\
\midrule
\multicolumn{4}{c}{\itshape ImageNet (ResNet34, 500 cls)} \\
\midrule
\method{} & \bfseries 1.000 & \bfseries 0.973 & 0.902 \\
FACE      & 0.998 & 0.952 & \bfseries 0.939 \\
CRAFT     & 0.976 & 0.918 & 0.808 \\
ICE       & 0.997 & 0.907 & 0.545 \\
\midrule
\multicolumn{4}{c}{\itshape Places365 (ResNet50, 364 cls)} \\
\midrule
\method{} & \bfseries 1.000 & \bfseries 0.972 & 0.880 \\
FACE      & 0.994 & 0.925 & \bfseries 0.919 \\
CRAFT     & 0.947 & 0.868 & 0.828 \\
ICE       & 0.995 & 0.887 & 0.631 \\
\bottomrule
\end{tabular}
\end{table}

\subsection{Predictive Preservation and Faithful Attribution}
\label{sec:main_results}

\Cref{tab:main_comparison} reports our primary comparison. All methods preserve
accuracy on both ResNet backbones, with \method{} highest, so the semantic
constraints do not distort the classifier's representations; with Acc saturated,
faithfulness is read from insertion and deletion. \method{} attains the highest
C-Ins on both datasets, so its named concepts are the most efficient at restoring
the prediction. On C-Del, FACE is strongest. This ordering reflects a design
difference, not a deficiency: FACE jointly optimizes coefficients and basis
against classifier logits, absorbing any decision-relevant direction including
non-interpretable ones, whereas \method{} fixes the coefficients to named concepts
and recovers a semantically constrained \emph{subset} of decision-relevant
evidence. Trading a portion of deletion completeness for named, verifiable
concepts is the intended behavior, and the gap narrows as the concept budget grows
(\cref{sec:ablation}).

\noindent
\textbf{Computational cost:}
\method{} is also cheaper to fit than the jointly-optimized baselines. Fixing
$\mathbf{S}$ reduces concept learning to a single non-negative least-squares solve
for $\mathbf{W}$, avoiding the alternating factor updates of CRAFT and the
logit-regularized optimization of FACE. On an RTX~A4500, \method{} fits a class in
$0.03$\,s, versus $0.5$\,s (CRAFT) and $2.6$\,s (FACE); a one-time LLM concept
generation step adds ${\sim}15$\,s/class. A full complexity and wall-clock
analysis is in the supplement.

\subsection{The Model's Geometry Determines Retained Concepts}
\label{sec:geometry}

\begin{figure*}[h]
    \centering
    \includegraphics[width=.9\textwidth]{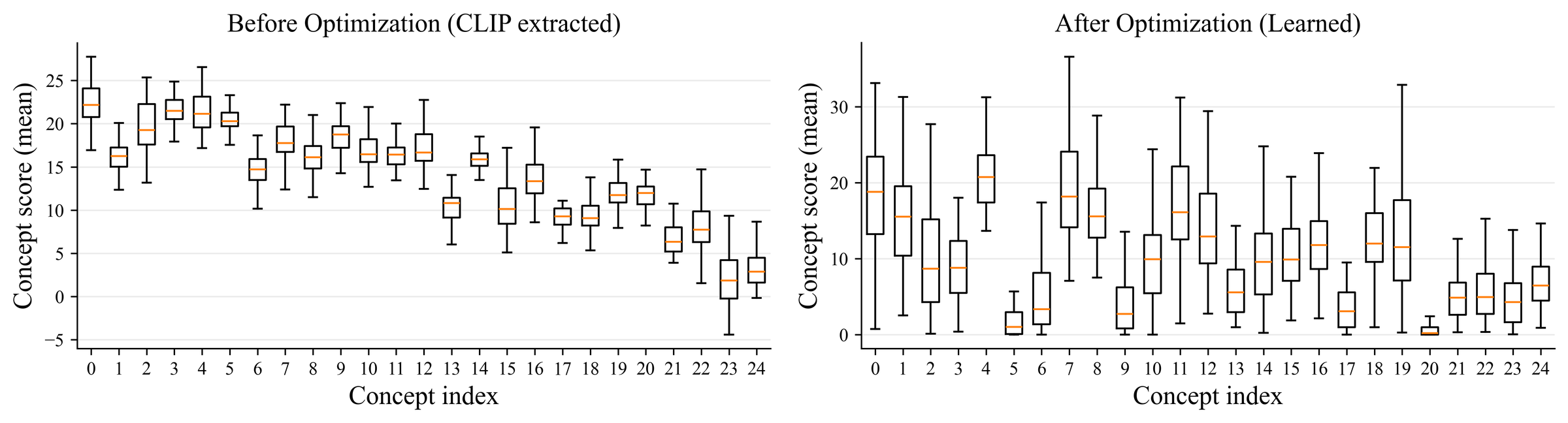}
   \caption{Concept activation scores before and after optimization, across 50
images. Before optimization, scores largely reflect the CLIP similarity
initialization; after, the reconstruction objective reshapes them to match encoder
activations, indicating the final concept maps are driven by the model's internal
representations rather than the initial CLIP similarity.}
    \label{fig:concept_score_before_after}
\end{figure*}

A central claim of \method{} is that the frozen encoder, not CLIP, determines
which named concepts survive. \Cref{fig:concept_score_before_after} provides
direct evidence: before optimization, concept scores reflect the CLIP
initialization and rank relatively uniformly; after optimizing the basis against
encoder activations, the distributions disperse as coefficients are reshaped to
reconstruct the model's internal features. Concepts inconsistent with the
encoder's evidence receive small basis weight and are suppressed, so \method{}
retains only concepts causally aligned with the decision, which matters given
known biases in CLIP-style models~\cite{tanjim2024discovering}.

\subsection{Concepts in a Clinical Domain}
\label{sec:retina}
\begin{figure}[t]
    \centering
    \includegraphics[width=\linewidth]{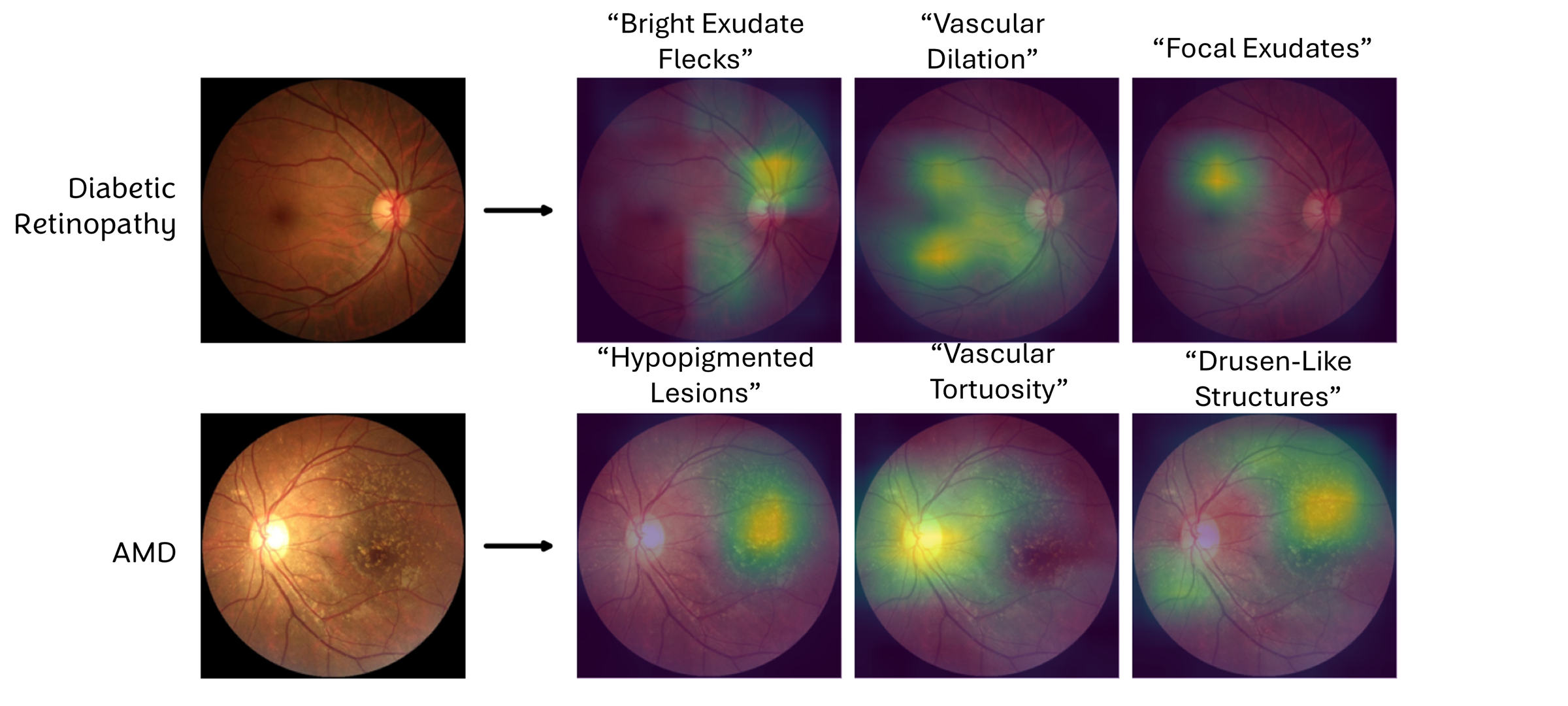}
    \caption{\textbf{Named-concept rollouts on retinal fundus images.} For a
    diabetic retinopathy case (top) and an age-related macular degeneration case
    (bottom), \method{} decomposes the prediction into separately named, localized
    concepts. Each map is anchored to a clinical term, so the explanation reports
    both where the model looks and what it attributes to that region. Two
    representative classes are shown; per-disease results are in the supplement.}
    \label{fig:retina_rollout}
\end{figure}

\noindent
To test whether \method{} produces decision-relevant named concepts where
interpretability carries real stakes, we evaluate on retinal fundus photographs
(ODIR-5K, five ocular-disease classes). The explained model is a DenseNet-121
fine-tuned to $0.755$ balanced accuracy; per-class vocabularies are generated as
disease-specific attributes (e.g., \emph{bright exudate flecks}, \emph{vascular
dilation}, \emph{drusen-like structures}) and localized with BiomedCLIP. Dataset
and training details are in the supplement.

\noindent
As on natural images, Acc and C-Ins saturate, so faithfulness is read from
deletion (C-Del), where \method{} is the most causally decisive of any method
(\cref{tab:retina}). Unlike the natural-image setting, where FACE's joint
optimization edges out \method{} on deletion, the clinical backbone's evidence is
spatially focal: disease signatures occupy small, localized regions rather than
distributed texture. \method{}'s named concepts anchor to these regions, so
removing them degrades the prediction more sharply, and \method{} leads C-Del by a
wide margin ($0.954$ vs.\ $0.863$). This aligns with our central finding that the
encoder's geometry, not the language prompts, determines which concepts are
retained (\cref{sec:geometry}): focal disease evidence aligns with named
anatomical concepts.

\noindent
\Cref{fig:retina_rollout} shows the qualitative counterpart. For diabetic
retinopathy, \method{} separates the prediction into named concepts localized to
distinct regions; for age-related macular degeneration, \emph{drusen-like
structures} localizes to the yellow deposits temporal to the optic disc, a
hallmark of the disease. Because each map carries a clinical name, a reader
inspects not only \emph{where} the model looks but \emph{what} it attributes
there, and can flag when the two disagree, making the rollout an auditable record
of the model's evidence rather than an unlabeled saliency map.

\begin{table}[t]
\centering
\footnotesize
\caption{Concept faithfulness on retinal fundus (DenseNet-121 @448, BiomedCLIP),
mean over the four disease classes (normal excluded). Acc and C-Ins saturate;
methods separate on C-Del, where \method{} leads by a wide margin. All metrics
$\uparrow$; best in bold. Per-disease results are in the supplement.}
\label{tab:retina}
\setlength{\tabcolsep}{18pt}
\begin{tabular}{l *{3}{S[table-format=1.3]}}
\toprule
Method & {Acc} & {C-Ins} & {C-Del} \\
\midrule
\method{} (ours) & \bfseries 1.000 & \bfseries 0.980 & \bfseries 0.954 \\
FACE             & 0.990 & 0.980 & 0.863 \\
CRAFT            & 1.000 & 0.980 & 0.850 \\
ICE              & 0.990 & 0.980 & 0.455 \\
\bottomrule
\end{tabular}
\end{table}

\subsection{Ablation Studies}
\label{sec:ablation}

\noindent
\textbf{The language anchor:}
\Cref{tab:language_ablation} replaces the fixed language-grounded coefficients
$\mathbf{S}$ with learned ones (standard unsupervised NMF on the same
$\bar{\mathbf{A}}$). Accuracy is unchanged, but removing the anchor collapses
deletion faithfulness, dropping C-Del from $0.902$ to $0.507$. The anchor changes
\emph{which} basis directions are discovered, steering them toward
classifier-meaningful structure that a purely data-driven factorization does not
recover.

\noindent
\textbf{Effect of concept budget $r$:}
\Cref{tab:rank_ablation} varies the number of concepts per class. Accuracy is
saturated at all $r$, while C-Ins and C-Del improve monotonically: a larger
vocabulary yields a more expressive basis that remains predictively faithful. We
fix $r=25$ in the main experiments to match the baselines' budgets; larger $r$
trades a modest gain in deletion faithfulness for reduced per-image
interpretability.

\begin{table}[t]
\centering
\footnotesize
\caption{The language anchor is functional, not cosmetic. Replacing fixed
language-grounded coefficients with learned ones leaves accuracy unchanged but
sharply reduces deletion faithfulness (C-Del). ImageNet, ResNet34, 500 classes,
$r{=}25$, identical splits. All metrics $\uparrow$.}
\label{tab:language_ablation}
\setlength{\tabcolsep}{8pt}
\begin{tabular}{l *{3}{S[table-format=1.3]}}
\toprule
Variant & {Acc} & {C-Ins} & {C-Del} \\
\midrule
\method{} (fixed $\mathbf{S}$) & \bfseries 1.000 & \bfseries 0.973 & \bfseries 0.902 \\
Unsupervised NMF on $\bar{\mathbf{A}}$ & 1.000 & 0.942 & 0.507 \\
\midrule
$\Delta$ (language) & {---} & {$+0.031$} & {$+0.395$} \\
\bottomrule
\end{tabular}
\end{table}

\begin{table}[t]
\centering
\footnotesize
\caption{Effect of concept budget $r$. Accuracy is saturated; faithfulness
improves monotonically with $r$. ImageNet, ResNet34, 500 classes.}
\label{tab:rank_ablation}
\setlength{\tabcolsep}{10pt}
\begin{tabular}{S[table-format=3.0] *{3}{S[table-format=1.3]} S[table-format=1.3]}
\toprule
{$r$} & {Acc $\uparrow$} & {C-Ins $\uparrow$} & {C-Del $\uparrow$} & {MSE $\downarrow$} \\
\midrule
10  & 1.000 & 0.941 & 0.860 & 1.477 \\
25  & 1.000 & 0.973 & 0.902 & 1.483 \\
50  & 1.000 & 0.985 & 0.953 & 1.543 \\
100 & 1.000 & 0.992 & 0.956 & 1.382 \\
\bottomrule
\end{tabular}
\end{table}

\section{Conclusion}

We introduced \method{}, a post-hoc framework that produces concepts which are
simultaneously named, faithful, and obtained without modifying the model. By
inverting non-negative matrix factorization, fixing CLIP-derived language-grounded
coefficients and learning only the concept basis against the frozen encoder,
\method{} ties each basis direction to a human-readable concept while preserving
the model's predictive behavior. Removing the language anchor preserves accuracy
but collapses faithfulness, showing that the anchor determines which concept
directions are discovered rather than merely labeling them.

\noindent Across ImageNet and Places365, \method{} matches the strongest prior method on
faithfulness while uniquely supporting per-image named-concept rollouts: for a
single input, the named concepts the model relied on, each localized and annotated
with its contribution. On retinal fundus images its named concepts are the most
causally decisive of any method, showing the approach carries to a high-stakes
clinical domain. By making a model's semantic evidence directly inspectable and
comparable across images, \method{} supports more transparent vision models where
understanding the content of a decision matters as much as its accuracy.
% \clearpage
% \input{sec/2_formatting}
% \input{sec/3_finalcopy}
{
    \small
    \bibliographystyle{ieeenat_fullname}
    \bibliography{main}
}
\clearpage
\maketitle
\appendix

\renewcommand{\thesection}{A\arabic{section}}
\renewcommand{\thesubsection}{A\arabic{section}.\arabic{subsection}}
\renewcommand{\thesubsubsection}{A\arabic{section}.\arabic{subsection}.\arabic{subsubsection}}

\renewcommand{\thefigure}{A\arabic{figure}}
\renewcommand{\thetable}{A\arabic{table}}
\renewcommand{\theequation}{A\arabic{equation}}

\setcounter{section}{0}
\setcounter{figure}{0}
\setcounter{table}{0}
\setcounter{equation}{0}

\begin{center}
{\Large \textbf{Supplementary Material}}
\end{center}

% ------------------------------------------------
% Organization
% ------------------------------------------------

\paragraph{Organization.}
This supplementary material provides additional implementation details,
theoretical analysis, and extended results supporting the main paper.
\begin{itemize}
    \item Sec.~\ref{app:concept_vocab} describes the construction of the
    class-specific concept vocabularies via LLM generation and semantic filtering,
    and the CLIP-based spatial probing used to build concept maps
    (Sec.~\ref{app:clip_maps}).
    \item Sec.~\ref{sec:supp_optimization} gives the optimization solver details
    deferred from the main paper.
    \item Sec.~\ref{apendix:convergence} provides the convergence analysis for the
    projected gradient descent procedure.
    \item Sec.~\ref{app:more_backbones} evaluates four additional ImageNet
    backbones, including a robustness analysis on a lightweight network.
    \item Sec.~\ref{app:dermoscopy} reports a second clinical domain (dermoscopy)
    and an analysis of where the discovered concepts localize.
    \item Sec.~\ref{app:compute} reports a per-stage computational cost and memory
    analysis against the baselines.
    \item Sec.~\ref{sec:supp_vit} presents a Vision Transformer case study.
    \item Sec.~\ref{apendix:concept_maps} shows extended qualitative
    visualizations of discovered concepts.
    \item Sec.~\ref{apendix:concept_comparison} provides additional qualitative
    comparisons with ICE, CRAFT, and FACE.
\end{itemize}

\section{Class-Specific Concept Vocabulary Generation}
\label{app:concept_vocab}

For each target class, we construct a fixed vocabulary of 25 visually grounded
concepts that serve as the semantic basis for our concept maps. The vocabulary is
generated using a constrained large language model (LLM) pipeline followed by
lexical and CLIP-based semantic filtering.

\subsection{Prompt Design}

Concept proposals are generated using the \texttt{gpt-4o-mini} model. The system
prompt instructs the model to generate short, visually grounded descriptors
suitable for interpretability analysis. The user prompt specifies the target
ImageNet class and enforces several constraints:

\begin{itemize}
\item Exactly 25 concepts must be generated.
\item Each concept must contain 2--3 lowercase words.
\item Concepts must describe visually grounded attributes such as parts, textures,
shapes, contexts, or actions.
\item The class name and trivial synonyms are disallowed.
\item Previously rejected or accepted concepts are added to a banned list to
encourage diversity across iterations.
\end{itemize}

Example prompt template:

\begin{verbatim}
Generate EXACTLY 25 distinct visual
concepts for the ImageNet class:
"{class_name}". Concepts must be 2-3
words, lowercase, visually grounded.
Avoid the class name or synonyms.
Avoid generic words such as "object",
"scene", "photo". Return JSON only.
\end{verbatim}

\subsection{Schema-Constrained Generation}

To ensure structured, reproducible outputs, generation is constrained by a strict
JSON schema: the output must contain the class name and exactly 25 concept strings,
\begin{equation}
\label{eq:concept_schema}
\texttt{schema} =
\left\{
\begin{array}{l}
\texttt{class\_name}: \text{string} \\
\texttt{concepts}: [c_1, \dots, c_{25}], \quad c_i \in \mathcal{S},
\end{array}
\right.
\end{equation}
where each concept $c_i$ is a 2--3 word lowercase phrase of 3--40 characters
matching the pattern
\begin{verbatim}
[a-z][a-z-]*([ ][a-z][a-z-]*){1,2}
\end{verbatim}
This guarantees the LLM returns exactly 25 short, controlled-format concepts.

\subsection{Stage-1 Lexical Filtering}

Raw LLM outputs are normalized through lowercasing and removal of non-alphabetic characters. A rule-based filter then removes concepts that violate the following conditions:

\begin{itemize}
\item Not 2--3 words long
\item Contain generic filler terms (e.g., \textit{animal, object, scene})
\item Duplicate previously accepted concepts
\item Contain tokens identical to the class name
\end{itemize}

To avoid removing meaningful attributes, we maintain a lexicon of visual attribute words spanning several categories:

\begin{itemize}
\item color (e.g., \textit{yellow, black})
\item texture (e.g., \textit{furry, shiny})
\item parts (e.g., \textit{tail, wings})
\item shape (e.g., \textit{long, round})
\item pose/motion (e.g., \textit{running, flying})
\item environment (e.g., \textit{forest, ocean})
\item material (e.g., \textit{metal, wood})
\end{itemize}

Concepts containing attribute words are preserved even if partial lexical overlap with the class name occurs.

\subsection{Stage-2 CLIP Semantic Filtering}

After lexical filtering, we apply a CLIP-based semantic filtering procedure to improve diversity and visual relevance.

Let \(t_i\) denote the CLIP text embedding of concept \(c_i\) obtained using the template:

\[
\text{``a photo of } c_i \text{''}.
\]

If class images are available, we randomly sample up to 100 images from the target class and compute their CLIP image embeddings. The mean embedding

\[
\mu_I = \frac{1}{N}\sum_{j=1}^{N} f_{\text{CLIP}}(I_j)
\]

serves as a prototype representation of the class.

Concepts are ranked by cosine similarity

\[
s_i = \langle t_i, \mu_I \rangle .
\]

To preserve semantic diversity, concepts are greedily selected while enforcing

\[
\max_{k < i} \langle t_i, t_k \rangle < 0.80 .
\]

This removes near-duplicate concepts in the CLIP embedding space while favoring those that align with the visual distribution of the class.

\subsection{Iterative Generation Procedure}

\Cref{alg:concept_generation} summarizes vocabulary generation. Early rounds apply
both lexical and CLIP-based filtering to encourage visual relevance and diversity.
If a class has not reached 25 concepts after the CLIP-filtered rounds, later
rounds disable CLIP pruning and keep only the lexical constraints, guaranteeing
exactly 25 valid concepts per class.

\begin{algorithm}[t]
\caption{Class-specific concept vocabulary generation}
\label{alg:concept_generation}
\small
\begin{algorithmic}[1]
\Require class name $c$, LLM $M$, CLIP encoders $f_{\text{text}}, f_{\text{img}}$
\Require optional class images $\mathcal{I}_c$; rounds $R_{\text{clip}}{=}10$, $R_{\text{fb}}{=}5$
\Ensure 25 filtered concepts $\mathcal{K}$
\State $\mathcal{K} \gets \emptyset$;\quad $\mathcal{B} \gets \emptyset$
   \Comment{accepted; banned}
\For{$r = 1$ to $R_{\text{clip}} + R_{\text{fb}}$}
    \State $\mathcal{C} \gets \textsc{LLMGenerate}(M, c, \mathcal{B})$
       \Comment{25 candidates}
    \State $\mathcal{C} \gets \textsc{LexicalFilter}(\mathcal{C}, c, \mathcal{K})$
    \If{$r \le R_{\text{clip}}$}
        \State $\mathcal{C} \gets \textsc{CLIPFilterRank}(\mathcal{C}, \mathcal{I}_c)$
           \Comment{skip in fallback}
    \EndIf
    \State $\mathcal{K} \gets \mathcal{K} \cup \mathcal{C}$;\quad
           $\mathcal{B} \gets \mathcal{K}$
    \If{$|\mathcal{K}| \ge 25$}
        \State \Return first 25 concepts of $\mathcal{K}$
    \EndIf
\EndFor
\State \textbf{raise error} \Comment{unreached in practice}
\end{algorithmic}
\end{algorithm}

\subsection{CLIP-Based Concept Map Construction}
\label{app:clip_maps}

After constructing the class-specific concept vocabulary (Sec.~\ref{app:concept_vocab}), we use CLIP to localize each concept within an image and build spatial concept maps. The goal is to measure how strongly each concept aligns with different spatial regions of the image.

\paragraph{CLIP model.}
We use a pretrained CLIP vision–language model to compute similarity between image regions and concept text prompts. In our implementation we use the ViT-B/16 architecture, loaded either through the \texttt{open\_clip} library with pretrained weights (e.g., \texttt{laion2b\_s34b\_b88k}) or the original OpenAI CLIP implementation when \texttt{open\_clip} is unavailable. All image and text features are $\ell_2$ normalized before computing cosine similarity.

\paragraph{Concept text embeddings.}
For each concept $c_i$ in the vocabulary of size $M=25$, we construct a CLIP text prompt using the template

\[
t_i = \text{``a photo of } c_i \text{''}.
\]

The CLIP text encoder produces a normalized embedding

\[
\mathbf{t}_i = f_{\text{text}}(t_i) \in \mathbb{R}^{D}.
\]

All text embeddings are precomputed and cached to avoid redundant computation during image processing.

\paragraph{Spatial probing via localized prompts.}
To obtain spatially localized responses, we probe the image using a grid of visual prompts. Each input image is first resized and center-cropped to the standard CLIP resolution of $224 \times 224$. We then define a uniform grid of $H_t \times W_t$ spatial locations (typically $7 \times 7$). For each grid cell center, we overlay a thin red circular outline with fixed radius $r$ on the image.

This produces a set of $H_t W_t$ modified images:

\[
\{ I_{p} \}_{p=1}^{H_t W_t},
\]

where each $I_p$ highlights a different spatial location through the circle prompt.

Each prompted image is encoded by the CLIP image encoder:

\[
\mathbf{v}_p = f_{\text{img}}(I_p) \in \mathbb{R}^{D}.
\]

The resulting features are also $\ell_2$ normalized.

\paragraph{Concept similarity maps.}
For each spatial location $p$ and concept $i$, we compute the cosine similarity between the image embedding and the concept embedding:

\[
S_{p,i} = \mathbf{v}_p^\top \mathbf{t}_i .
\]

Optionally, the similarity scores are scaled by the CLIP temperature parameter $\exp(\text{logit\_scale})$.

The resulting similarity tensor

\[
S \in \mathbb{R}^{H_t W_t \times M}
\]

is reshaped into a spatial concept map

\[
P_i \in \mathbb{R}^{H_t \times W_t}
\]

for each concept $i$. The stack of maps

\[
P = \{P_i\}_{i=1}^{M}
\]

represents the spatial activation of all concepts across the image.

% \paragraph{Concept importance estimation.}
% To estimate the importance of each concept within an image, we compute a binary activation mask for each map using Otsu thresholding. For concept map $P_i$, the threshold $\tau_i$ is computed using Otsu's method on the map values. The resulting mask is

% \[
% M_i(x,y) = \mathbf{1}[P_i(x,y) \ge \tau_i].
% \]

% The concept importance score is defined as the average activation within the selected region:

% \[
% \text{importance}_i =
% \frac{\sum_{x,y} P_i(x,y) M_i(x,y)}
% {\sum_{x,y} M_i(x,y) + \epsilon}.
% \]

% \paragraph{Stored concept map representation.}
% For each processed image we store the following data:

% \begin{itemize}
% \item concept activation maps $P \in \mathbb{R}^{M \times H_t \times W_t}$
% \item concept importance scores
% \item Otsu activation masks
% \item concept vocabulary
% \item spatial grid metadata (centers and grid size)
% \end{itemize}

% These concept maps serve as the semantic representation used in the subsequent concept factorization and explanation stages of our method.
\section{Optimization Details}
\label{sec:supp_optimization}

This section gives the solver details for the basis-learning objective
(Eq.~1) and the inference problem (Eq.~3)
in the main paper.

\noindent
\textbf{Basis update.} We enforce non-negativity in Eq.~1 with
the projected gradient step
\begin{equation}
\mathbf{W} \leftarrow \max\!\big(0,\ \mathbf{W} - \eta \nabla_{\mathbf{W}}\mathcal{L}_{\text{recon}}\big),
\label{eq:supp_pgd_w}
\end{equation}
initialized with NNDSVD~\cite{boutsidis2008svd} for stable, accelerated
convergence. With $\mathbf{S}$ fixed, $\mathcal{L}_{\text{recon}}$ is a smooth
quadratic in $\mathbf{W}$, so projected gradient descent converges to a stationary
point.

\noindent
\textbf{Inference solver.} For the inference problem
(Eq.~3), a fast non-negative initialization is the projected
normal equation
\begin{equation}
\hat{\mathbf{S}}_0 =
\operatorname{ReLU}\!\left(\bar{\mathbf{A}}\mathbf{W}(\mathbf{W}^{\top}\mathbf{W})^{-1}\right),
\label{eq:supp_normal_eq}
\end{equation}
optionally refined by projected gradient descent,
\begin{equation}
\hat{\mathbf{S}} \leftarrow
\operatorname{ReLU}\!\left(\hat{\mathbf{S}} - \eta(\hat{\mathbf{S}}\mathbf{W}^{\top}\mathbf{W} - \bar{\mathbf{A}}\mathbf{W})\right),
\label{eq:supp_pgd_s}
\end{equation}
where the step size $\eta$ is set from the spectral norm of
$\mathbf{W}^{\top}\mathbf{W}$.

\noindent
\textbf{Heatmap normalization.} For visualization, each concept channel is
rescaled to $[0,1]$,
\begin{equation}
\mathbf{M}_k = \mathrm{Normalize}\!\left(\hat{\mathbf{S}}_{\text{spatial}}(:,:,k)\right).
\label{eq:supp_heatmap}
\end{equation}
% ============================================================
\section{PGD Convergence Analysis}
\label{apendix:convergence}

We analyze the convergence of the projected gradient descent (PGD)
procedure used to optimize the activation-level reconstruction objective
introduced in Sec.~3.4 of the main paper. In our formulation the semantic
coefficient matrix $\mathbf{S}$ is fixed, and we learn only the
nonnegative concept basis $\mathbf{W}$.

Let $\bar{\mathbf{A}} \in \mathbb{R}^{N \times p}$ denote the unfolded
encoder activations, where $N = n h w$ corresponds to the number of
spatial locations across all images. The matrix
$\mathbf{S} \in \mathbb{R}^{N \times r}$ represents the fixed
CLIP-derived semantic similarity matrix. The optimization problem for
the concept basis $\mathbf{W} \in \mathbb{R}^{p \times r}_{+}$ is

\begin{equation}
\label{eq:opt_recon}
\min_{\mathbf{W} \ge 0}
\mathcal{L}(\mathbf{W})
=
\frac{1}{2}
\|\bar{\mathbf{A}} - \mathbf{S}\mathbf{W}^\top\|_F^2 .
\end{equation}

This objective enforces activation-level fidelity between the original
latent activations $\bar{\mathbf{A}}$ and the reconstruction
$\mathbf{S}\mathbf{W}^\top$ obtained from the concept coefficients
$\mathbf{S}$ and concept bases $\mathbf{W}$.

\paragraph{Smoothness.}
The loss in Eq.~\ref{eq:opt_recon} is a convex quadratic function of
$\mathbf{W}$. Rewriting the objective as

\[
\mathcal{L}(\mathbf{W})
=
\frac{1}{2}
\|\bar{\mathbf{A}}^\top - \mathbf{W}\mathbf{S}^\top\|_F^2 ,
\]

its gradient with respect to $\mathbf{W}$ is

\begin{equation}
\nabla_{\mathbf{W}} \mathcal{L}(\mathbf{W})
=
\mathbf{W}(\mathbf{S}^\top \mathbf{S})
-
\bar{\mathbf{A}}^\top \mathbf{S}.
\end{equation}

Since $\mathbf{S}^\top\mathbf{S}$ is positive semidefinite, the gradient
is Lipschitz continuous with constant

\begin{equation}
L_\nabla = \|\mathbf{S}^\top\mathbf{S}\|_2 .
\end{equation}

Therefore the objective $\mathcal{L}(\mathbf{W})$ is $L_\nabla$–smooth.

\paragraph{Projected gradient update.}
To enforce the nonnegativity constraint on $\mathbf{W}$, we apply
projected gradient descent. The update rule is

\begin{equation}
\label{eq:pgd_update}
\mathbf{W}_{k+1}
=
\Pi_{\mathbb{R}_+^{p\times r}}
\left(
\mathbf{W}_k
-
\eta
\nabla \mathcal{L}(\mathbf{W}_k)
\right),
\end{equation}

where $\Pi_{\mathbb{R}_+^{p\times r}}(\cdot)$ denotes the Euclidean
projection onto the nonnegative orthant, implemented elementwise as
$\max(0,\cdot)$.

\paragraph{Descent property.}
For an $L_\nabla$–smooth function, the standard descent lemma gives

\begin{equation}
\label{eq:descent}
\mathcal{L}(\mathbf{W}')
\le
\mathcal{L}(\mathbf{W})
+
\langle
\nabla \mathcal{L}(\mathbf{W}),
\mathbf{W}'-\mathbf{W}
\rangle
+
\frac{L_\nabla}{2}
\|\mathbf{W}'-\mathbf{W}\|_F^2 .
\end{equation}

Applying Eq.~\ref{eq:descent} with $\mathbf{W}' = \mathbf{W}_{k+1}$
and using the optimality condition of the projection step yields

\begin{equation}
\langle
\nabla \mathcal{L}(\mathbf{W}_k),
\mathbf{W}_{k+1}-\mathbf{W}_k
\rangle
\le
-\frac{1}{\eta}
\|\mathbf{W}_{k+1}-\mathbf{W}_k\|_F^2 .
\end{equation}

Substituting this bound into Eq.~\ref{eq:descent} gives

\begin{equation}
\mathcal{L}(\mathbf{W}_{k+1})
\le
\mathcal{L}(\mathbf{W}_k)
-
\left(
\frac{1}{\eta}
-
\frac{L_\nabla}{2}
\right)
\|\mathbf{W}_{k+1}-\mathbf{W}_k\|_F^2 .
\end{equation}

Therefore, for any step size satisfying

\begin{equation}
0 < \eta < \frac{2}{L_\nabla},
\end{equation}

the objective decreases monotonically,

\[
\mathcal{L}(\mathbf{W}_{k+1}) \le \mathcal{L}(\mathbf{W}_k).
\]

\paragraph{Convergence.}
Because $\mathcal{L}(\mathbf{W}) \ge 0$, the sequence
$\{\mathcal{L}(\mathbf{W}_k)\}$ is monotonically decreasing and
therefore convergent. Furthermore,

\[
\|\mathbf{W}_{k+1}-\mathbf{W}_k\|_F \rightarrow 0 .
\]

Every accumulation point of the PGD iterates satisfies the
first-order optimality condition for Eq.~\ref{eq:opt_recon}. Since the
objective is convex and the feasible set
$\{\mathbf{W} \ge 0\}$ is convex, any such stationary point is a
global minimizer of the problem.
% (your existing convergence section here)

% ============================================================
\section{Additional Backbones}
\label{app:more_backbones}

To test whether \method{}'s behavior is consistent across encoder architectures,
we evaluate four additional ImageNet backbones spanning different design families:
RegNetY-8GF, ResNet-152, ResNeXt-50 (32$\times$4d), and MobileNetV3-Large
(\cref{tab:more_backbones}). All methods use identical preprocessing, data splits,
and concept count ($r{=}25$).

\paragraph{Consistency with the main results.}
The pattern matches the main paper. \method{} preserves reconstruction accuracy
(Acc $\ge 0.999$ on all four) and attains the highest or tied concept insertion
(C-Ins) on every backbone, so its named concepts remain the most efficient at
restoring the prediction. On deletion (C-Del), \method{} and FACE trade the lead,
as on ImageNet and Places365: \method{} leads on RegNetY-8GF, while FACE edges it
on ResNet-152 and ResNeXt-50. The gap is small in every case, consistent with
\method{} recovering a named, decision-relevant subset of the evidence FACE
absorbs through unconstrained logit alignment.

\paragraph{Robustness on a lightweight backbone.}
MobileNetV3-Large is the most informative case. \method{} remains stable
(Acc $0.999$, C-Ins $0.962$, C-Del $0.946$), whereas CRAFT and FACE degrade
sharply and become highly unstable across classes
(C-Del $0.804{\pm}0.25$ and $0.781{\pm}0.22$ versus \method{}'s $0.946$). Fixing
the coefficients to language-grounded concepts evidently regularizes the
decomposition on compact, low-redundancy architectures where the baselines'
jointly-optimized factors do not reliably converge.

\begin{table}[t]
\centering
\footnotesize
\caption{\method{} across four additional ImageNet backbones ($r{=}25$, identical
splits). \method{} leads or ties Acc and C-Ins on all four; \method{} and FACE
trade the deletion lead, as in the main paper. On MobileNetV3-Large the baselines
become unstable (large per-class std), while \method{} remains stable. Best C-Del
per backbone in bold. All metrics $\uparrow$.}
\label{tab:more_backbones}
\setlength{\tabcolsep}{5pt}
\begin{tabular}{ll *{3}{S[table-format=1.3]}}
\toprule
Backbone & Method & {Acc} & {C-Ins} & {C-Del} \\
\midrule
\multirow{3}{*}{RegNetY-8GF}
 & \method{} & 1.000 & 0.977 & \bfseries 0.953 \\
 & CRAFT     & 0.997 & 0.971 & 0.940 \\
 & FACE      & 0.998 & 0.977 & 0.951 \\
\midrule
\multirow{3}{*}{ResNet-152}
 & \method{} & 0.999 & 0.975 & 0.914 \\
 & CRAFT     & 0.995 & 0.971 & 0.867 \\
 & FACE      & 0.995 & 0.975 & \bfseries 0.952 \\
\midrule
\multirow{3}{*}{ResNeXt-50 (32$\times$4d)}
 & \method{} & 1.000 & 0.972 & 0.867 \\
 & CRAFT     & 0.992 & 0.963 & 0.861 \\
 & FACE      & 0.992 & 0.974 & \bfseries 0.961 \\
\midrule
\multirow{3}{*}{MobileNetV3-Large}
 & \method{} & 0.999 & 0.962 & \bfseries 0.946 \\
 & CRAFT     & 0.839 & 0.805 & 0.804 \\
 & FACE      & 0.856 & 0.813 & 0.781 \\
\bottomrule
\end{tabular}
\end{table}

% ============================================================

\section{Second Clinical Domain: Dermoscopy}
\label{app:dermoscopy}

To test whether \method{}'s clinical results extend beyond retinal fundus, we add a
second medical domain, skin-lesion dermoscopy (HAM10000), explained with the same
domain-matched medical CLIP (BiomedCLIP) used for fundus.

\paragraph{Setup.}
HAM10000 contains $10{,}015$ dermoscopy images across seven lesion classes, split
at the lesion level (all shots of a lesion kept together to prevent leakage) into
$8{,}021$ train / $1{,}994$ validation (seed $42$). The explained model is a
ConvNeXt-Small with an ImageNet-pretrained encoder and a $768\!\rightarrow\!7$
linear head, trained at $224$ then fine-tuned at $448$ with
inverse-frequency-weighted cross-entropy and selected by balanced accuracy,
reaching $0.846$ balanced accuracy ($0.878$ top-1). Concepts are generated as in
Sec.~\ref{app:concept_vocab} (25/class, \texttt{gpt-4o-mini}) and localized with
BiomedCLIP by red-circle prompting; full setup is in \cref{tab:med_setup}.

\paragraph{Results.}
\Cref{tab:dermoscopy} reports concept faithfulness on both clinical domains under
identical metric code and a matched $7{\times}7$ probe grid, so differences reflect
the concept basis rather than the evaluation. As on retinal fundus, accuracy
saturates and \method{} leads concept deletion (C-Del), exceeding the strongest
baseline on dermoscopy ($0.418$ vs.\ CRAFT $0.366$); \method{} also retains
saturated insertion ($0.980$) whereas FACE's drops to $0.790$. Absolute C-Del,
however, is far lower than on fundus ($0.418$ vs.\ $0.954$) for every method.

\begin{figure}[t]
    \centering
    \includegraphics[width=\linewidth]{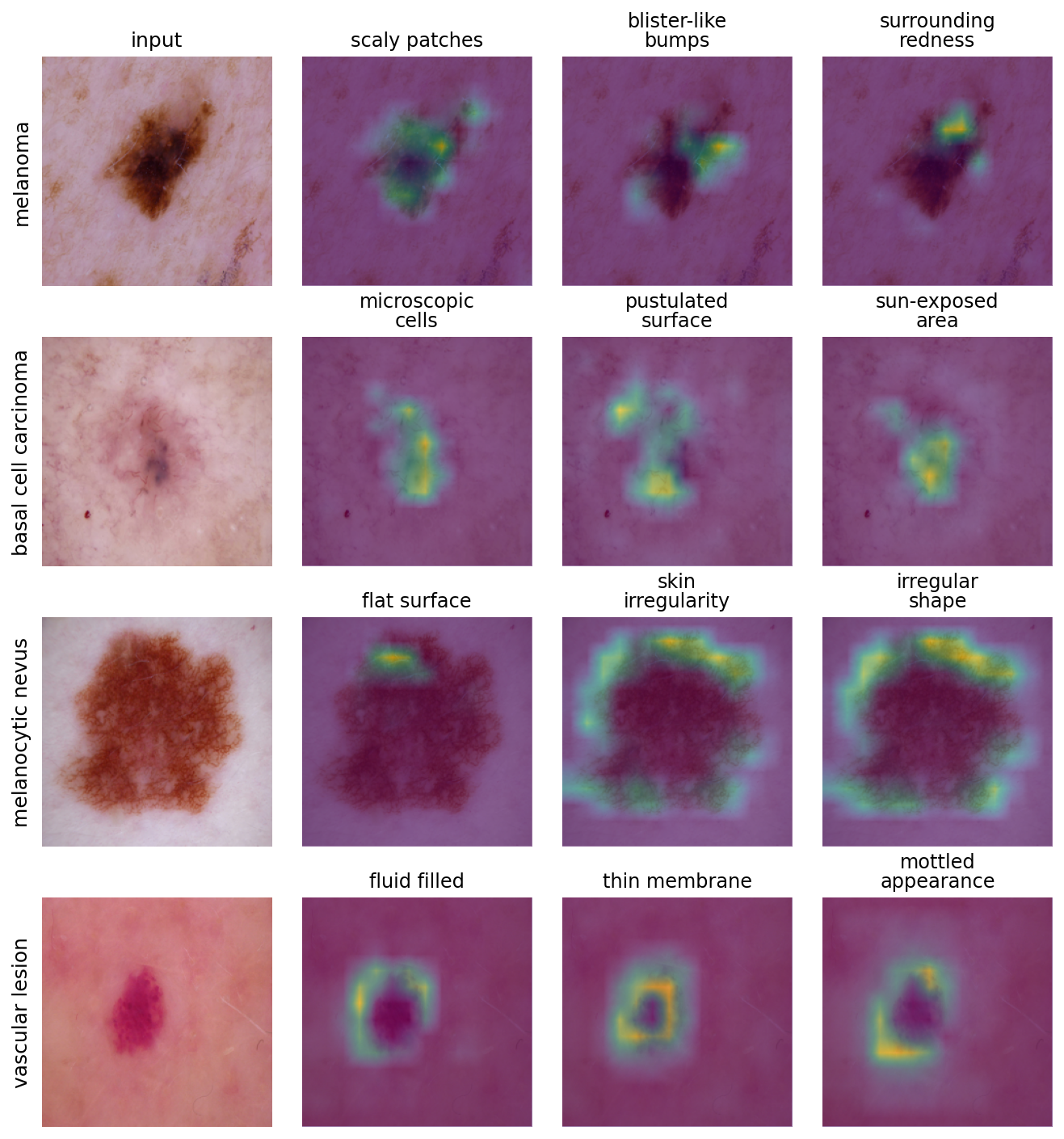}
    \caption{\textbf{\method{} concepts on dermoscopy (HAM10000, BiomedCLIP).}
    Top concepts for four lesion classes. The maps are spatially coherent and
    consistent but localize mainly to the lesion \emph{boundary} and surrounding
    skin (e.g., ``irregular shape'' and ``skin irregularity'' ringing the nevus;
    ``fluid filled''/``thin membrane'' encircling the vascular lesion) rather than
    the diagnostic interior, with basal cell carcinoma a partial exception. This
    boundary bias is consistent with the lower concept-deletion score in this
    domain (\cref{tab:dermoscopy}).}
    \label{fig:derm_concepts}
\end{figure}

\paragraph{Where the concepts localize.}
\Cref{fig:derm_concepts} explains the lower deletion score. The named concepts
localize predominantly to the lesion \emph{border} and peri-lesional skin rather
than to the diagnostic lesion \emph{interior}. For melanocytic nevus, ``skin
irregularity'' and ``irregular shape'' form a ring around the lesion edge while
the pigmented center stays inactive; for vascular lesion, ``fluid filled,''
``thin membrane,'' and ``mottled appearance'' encircle the rim; for melanoma,
``surrounding redness'' and ``scaly patches'' fall on the margin and adjacent
skin. Basal cell carcinoma is the main exception, with concepts landing on the
lesion body. This boundary bias and the low C-Del are the same phenomenon:
because the discriminative interior texture (pigment network, color variegation)
is largely untouched, removing the boundary-localized concepts leaves the evidence
the head relies on intact, so the prediction changes little. We report this as a
limitation: in this domain \method{}'s named concepts are spatially coherent and
consistent across images, but they often track lesion shape and context rather
than the internal morphology a dermatologist weighs most, so the explanations are
better read as localizing \emph{where} the model's lesion-level evidence sits than
as asserting clinically complete reasoning.

\begin{table}[t]
\centering
\footnotesize
\caption{Concept faithfulness across two clinical domains (both BiomedCLIP,
$r{=}25$) under identical metric code. \method{} leads C-Del in both. Dermoscopy
uses a matched $14{\times}14$ probe grid for all methods; absolute C-Del is lower on
dermoscopy, where the concepts localize to the lesion boundary rather than its
interior (\cref{fig:derm_concepts}). All metrics $\uparrow$; best C-Del per domain
in bold.}
\label{tab:dermoscopy}
\setlength{\tabcolsep}{5pt}
\begin{tabular}{ll *{3}{S[table-format=1.3]}}
\toprule
Domain & Method & {Acc} & {C-Ins} & {C-Del} \\
\midrule
\multirow{4}{1.5cm}{Retinal fundus}
 & \method{} & 1.000 & 0.980 & \bfseries 0.954 \\
 & FACE      & 0.990 & 0.980 & 0.863 \\
 & CRAFT     & 1.000 & 0.980 & 0.850 \\
 & ICE       & 0.990 & 0.980 & 0.455 \\
\midrule
\multirow{3}{1.5cm}{Dermoscopy}
 & \method{} & 1.000 & 0.980 & \bfseries 0.418 \\
 & CRAFT     & 0.980 & 0.940 & 0.366 \\
 & FACE      & 0.990 & 0.790 & 0.254 \\
\bottomrule
\end{tabular}
\end{table}

\begin{table}[t]
\centering
\footnotesize
\caption{Dataset, backbone, and concept-discovery setup for the two clinical
domains. Both backbones are ImageNet-pretrained, head-replaced, trained at $224$
then fine-tuned at $448$ with weighted cross-entropy and balanced-accuracy
selection; both use BiomedCLIP for concept localization.}
\label{tab:med_setup}
\begin{tabular}{lll}
\toprule
 & Retinal & Dermoscopy \\
\midrule
Dataset        & ODIR-5K            & HAM10000 \\
Classes        & 5                  & 7 \\
Train / Val    & 3{,}376 / 843      & 8{,}021 / 1{,}994 \\
Backbone       & DenseNet-121       & ConvNeXt-Small \\
CLIP           & BiomedCLIP         & BiomedCLIP \\
Probe grid     & $14{\times}14$     & $14{\times}14$ \\
PGD steps      & 250                & 150 \\
Bal. acc.      & 0.755              & 0.846 \\
\bottomrule
\end{tabular}
\end{table}
\section{Computational Cost}
\label{app:compute}

\Cref{tab:compute_stages} breaks down the per-class cost of \method{} on an
RTX~A4500 (ResNet34, $r{=}25$). The recurring concept-learning cost, the basis
fit, is $0.03$\,s, an order of magnitude below CRAFT ($0.5$\,s) and FACE
($2.6$\,s), because fixing $\mathbf{S}$ reduces basis learning to a single
non-negative least-squares solve rather than alternating factor updates. Inference
is effectively free. The dominant cost is CLIP concept-map construction, a
one-time, fully cacheable preprocessing step; the LLM vocabulary is generated once
($\sim$15\,s/class) and cached.

\begin{table}[h]
\centering
\footnotesize
\caption{Per-class cost on an RTX~A4500 (ResNet34, $r{=}25$). The recurring fit is
the concept-basis solve, where \method{} is far faster than the baselines. CLIP
map construction and LLM vocabulary are one-time, cacheable preprocessing.
Memory is peak GPU during each stage.}
\label{tab:compute_stages}
\begin{tabular}{l S[table-format=1.2] S[table-format=4.0]}
\toprule
Stage & {Wall (s)} & {GPU peak (MB)} \\
\midrule
Activation extraction        & 0.01 & 926 \\
CLIP map construction        & 6.45 & 1729 \\
LLM vocab (one-time)         & 15.0 & 0 \\
\midrule
\method{} fit                & \bfseries 0.03 & 745 \\
\method{} inference          & {$<0.01$} & 722 \\
CRAFT fit                    & 0.50 & 783 \\
FACE fit                     & 2.60 & 795 \\
\bottomrule
\end{tabular}
\end{table}

\paragraph{CLIP map construction is the bottleneck, and it is bounded.}
Concept-map construction probes a $14{\times}14$ grid of red-circle prompts per
image, so its cost scales with the number of CLIP forward passes, not with the
decomposition. It is also the only memory-significant stage. \Cref{tab:compute_clip}
shows peak GPU memory as a function of CLIP batch size: the whole pipeline fits in
under $1.8$\,GB even at the default batch, with throughput around $1{,}000$
prompted variants per second. The step is parallel across grid
locations and images and is computed once per class, then cached.

\begin{table}[h]
\centering
\footnotesize
\caption{CLIP map construction: peak GPU memory and throughput versus batch size
($14{\times}14$ grid, RTX~A4500). Memory grows gracefully with batch; the pipeline
default (196, one full grid per image) peaks under $1.8$\,GB.}
\label{tab:compute_clip}
\begin{tabular}{S[table-format=3.0] S[table-format=4.0] S[table-format=4.0]}
\toprule
{CLIP batch} & {GPU peak (MB)} & {Variants/s} \\
\midrule
16  & 976  & 1158 \\
32  & 1045 & 1156 \\
64  & 1183 & 988 \\
128 & 1450 & 1032 \\
196 & 1729 & 970 \\
\bottomrule
\end{tabular}
\end{table}

\paragraph{Grid sensitivity.}
The probe grid sets both the spatial resolution of the concept maps and the cost.
\Cref{tab:compute_grid} compares the $7{\times}7$ grid (used for ImageNet and
dermoscopy) with the $14{\times}14$ grid (used for the retinal fundus models). The
finer grid quadruples the prompted variants per image ($49\to196$) and costs
roughly $3.4\times$ the time and $1.6\times$ the peak memory, the price paid for
the higher-resolution localization that focal retinal evidence benefits from.

\begin{table}[h]
\centering
\footnotesize
\caption{Grid sensitivity of CLIP map construction (32 images, RTX~A4500). The
$14{\times}14$ grid used for retinal fundus costs more than the $7{\times}7$ grid
but yields higher-resolution concept maps.}
\label{tab:compute_grid}
\begin{tabular}{l S[table-format=3.0] S[table-format=1.2] S[table-format=4.0]}
\toprule
Grid & {Variants/img} & {Wall (s, 32 img)} & {GPU peak (MB)} \\
\midrule
$7{\times}7$   & 49  & 1.88 & 1116 \\
$14{\times}14$ & 196 & 6.37 & 1737 \\
\bottomrule
\end{tabular}
\end{table}
\section{Vision Transformer Case Study}
\label{sec:supp_vit}

NMF-style concept discovery (ICE, CRAFT, FACE) is formulated for convolutional
feature maps: it factorizes a spatial activation tensor whose channels share a
consistent, translation-equivariant geometry. Vision Transformers break both
assumptions, they emit patch tokens rather than a spatial map, mix global
information through self-attention at every layer, and route the prediction
through a \texttt{[CLS]} token rather than pooled spatial features. Consequently
no existing post-hoc concept-discovery method is designed to operate on both CNNs
and ViTs, and the convolutional baselines do not apply to patch tokens at all.
Because \method{} fixes language-grounded coefficients and learns only a basis,
it can be \emph{run} on ViT patch tokens without modification, which lets us ask a
question the field has not examined: when an anchored decomposition is applied to
a transformer, \emph{what kind of concepts does it actually recover?}

\paragraph{Setup.}
We decompose the penultimate patch-token representation of ViT-B/16
($14{\times}14{\times}768$) into the language-anchored basis and let the remaining
block, LayerNorm, and head complete the forward pass, on the same 500 ImageNet
classes ($r{=}25$).

\paragraph{Reconstruction transfers, but localization does not.}
Numerically, the decomposition reconstructs the representation well:
reconstruction accuracy and concept insertion are saturated (Acc $=0.98$,
C-Ins $=0.98$). Concept deletion, however, is near zero (C-Del $=0.02$). Two
factors combine here. First, the head reads from the preserved \texttt{[CLS]}
token, which carries information from the original forward pass, so ablating
patch-token concepts has little effect on the logits, and CNN-style deletion is
not directly meaningful. Second, and more telling, the recovered concept maps are
weakly localized.

\begin{figure}[t]
    \centering
    \includegraphics[width=\linewidth]{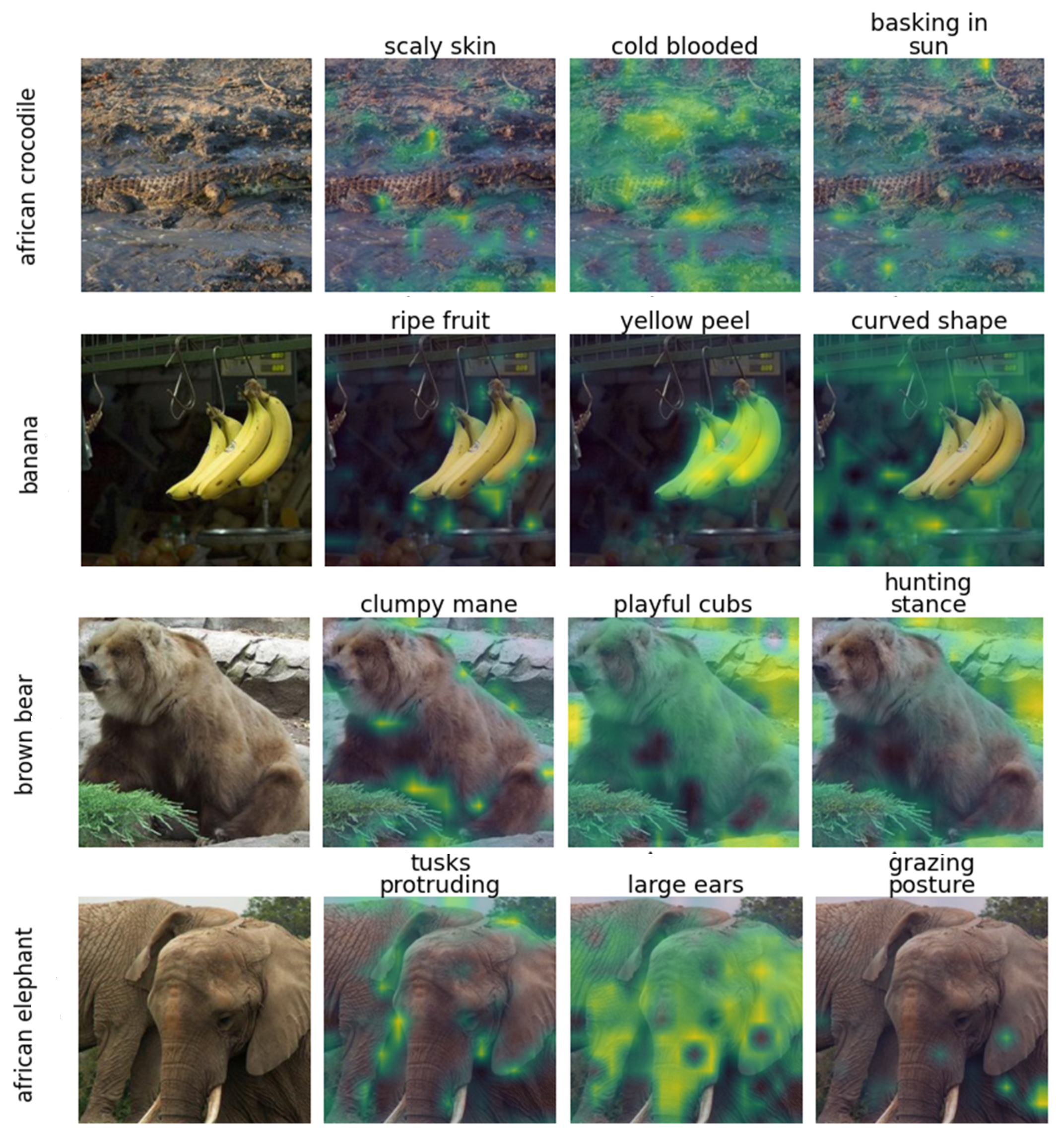}
    \caption{\textbf{\method{} concepts on ViT-B/16.} For four ImageNet classes,
    the top language-anchored concepts on patch tokens. Unlike the part-level
    localization \method{} produces on CNNs (Sec.~5.1, main paper), the ViT maps
    are diffuse and often attach to background or to the object silhouette rather
    than the named part (e.g., ``cold blooded'' on surrounding rock, ``large ears''
    over the whole head). Color- and texture-defined concepts localize better
    (``yellow peel''). The maps behave closer to global saliency than to
    part-grounded evidence, illustrating why spatial concept decomposition does not
    transfer directly to transformers.}
    \label{fig:vit_concepts}
\end{figure}

\Cref{fig:vit_concepts} makes this visible. On CNNs, \method{} localizes named
concepts to the corresponding object part (Sec.~5.1, main paper); on ViT, the same
concepts spread diffusely and frequently attach to the \emph{background} or trace
the object \emph{silhouette} rather than the named part. ``Cold blooded'' for the
crocodile activates on the surrounding rock and water; ``large ears'' on the
elephant covers the whole head and contour rather than the ears; the bear's
``hunting stance'' and ``playful cubs'' fire across the entire frame. A few
color- or texture-defined concepts do localize (``yellow peel'' and ``ripe fruit''
on the banana), but the part- and pose-level concepts that localize cleanly on
CNNs do not. The maps behave closer to global saliency or edge responses than to
the part-grounded evidence \method{} recovers on convolutional encoders.

\paragraph{Takeaway.}
This is a negative result, and an informative one: it shows \emph{why} spatial
concept decomposition is not readily transferable to transformers. The patch-token
geometry, global attention mixing, and \texttt{[CLS]}-routed prediction together
mean that a language-anchored basis can reconstruct the representation
(high insertion) without its concepts being spatially faithful (low deletion,
diffuse maps). A transformer-native formulation, operating on the \texttt{[CLS]}
attention pathway rather than patch tokens, and a \texttt{[CLS]}-aware deletion
protocol, are needed before concept discovery is meaningful on ViTs; we view this
as the natural next step rather than a property of \method{} specifically.

\begin{table}[h]
\centering
\footnotesize
\caption{\method{} on ViT-B/16, 500 ImageNet classes, $r{=}25$. The anchored
decomposition reconstructs the representation (Acc, C-Ins saturated) but its
concepts are not spatially faithful: deletion is near zero and the maps are
diffuse (\cref{fig:vit_concepts}). C-Del is not comparable to the CNN setting
under the preserved \texttt{[CLS]} pathway.}
\label{tab:vit}
\begin{tabular}{l *{3}{S[table-format=1.2]}}
\toprule
Backbone & {Acc $\uparrow$} & {C-Ins $\uparrow$} & {C-Del} \\
\midrule
ViT-B/16 & 0.98 & 0.98 & 0.02$^{\dagger}$ \\
\bottomrule
\end{tabular}
\\[2pt]
{\footnotesize $\dagger$ Not comparable to CNNs (preserved \texttt{[CLS]} pathway;
see text).}
\end{table}

% ============================================================
\section{Visual analysis of  concept maps}\label{apendix:concept_maps}
\begin{figure*}[t!]
    \centering
    \includegraphics[width= .9\textwidth ]{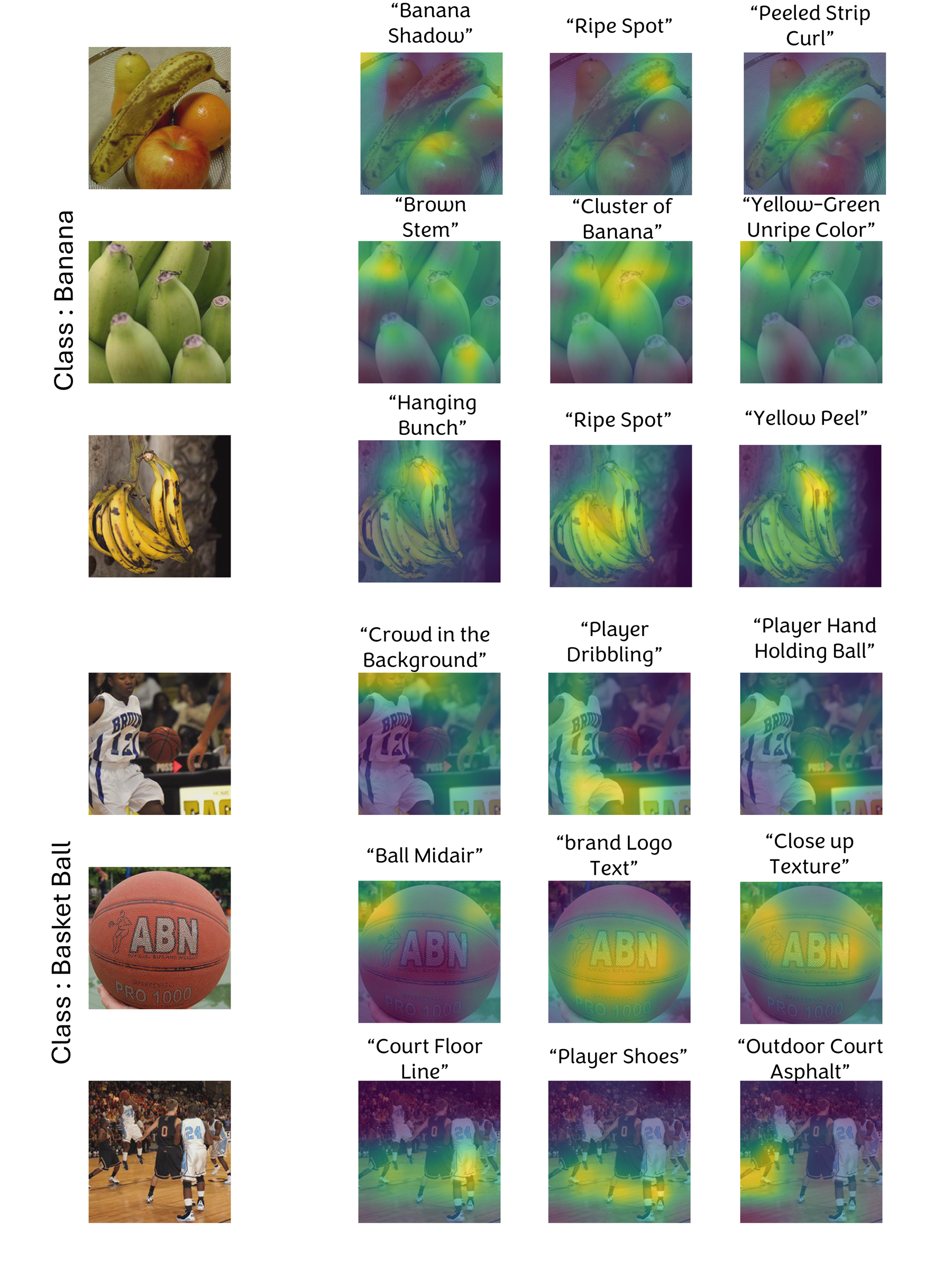}
   \caption{
Concept maps for the Banana and Basketball classes, with the highest-activating learned concepts overlaid on input images.
}
    \label{fig:qualitative_ours_banana_basketball}
\end{figure*}

\begin{figure*}[t!]
    \centering
    \includegraphics[width= .9\textwidth ]{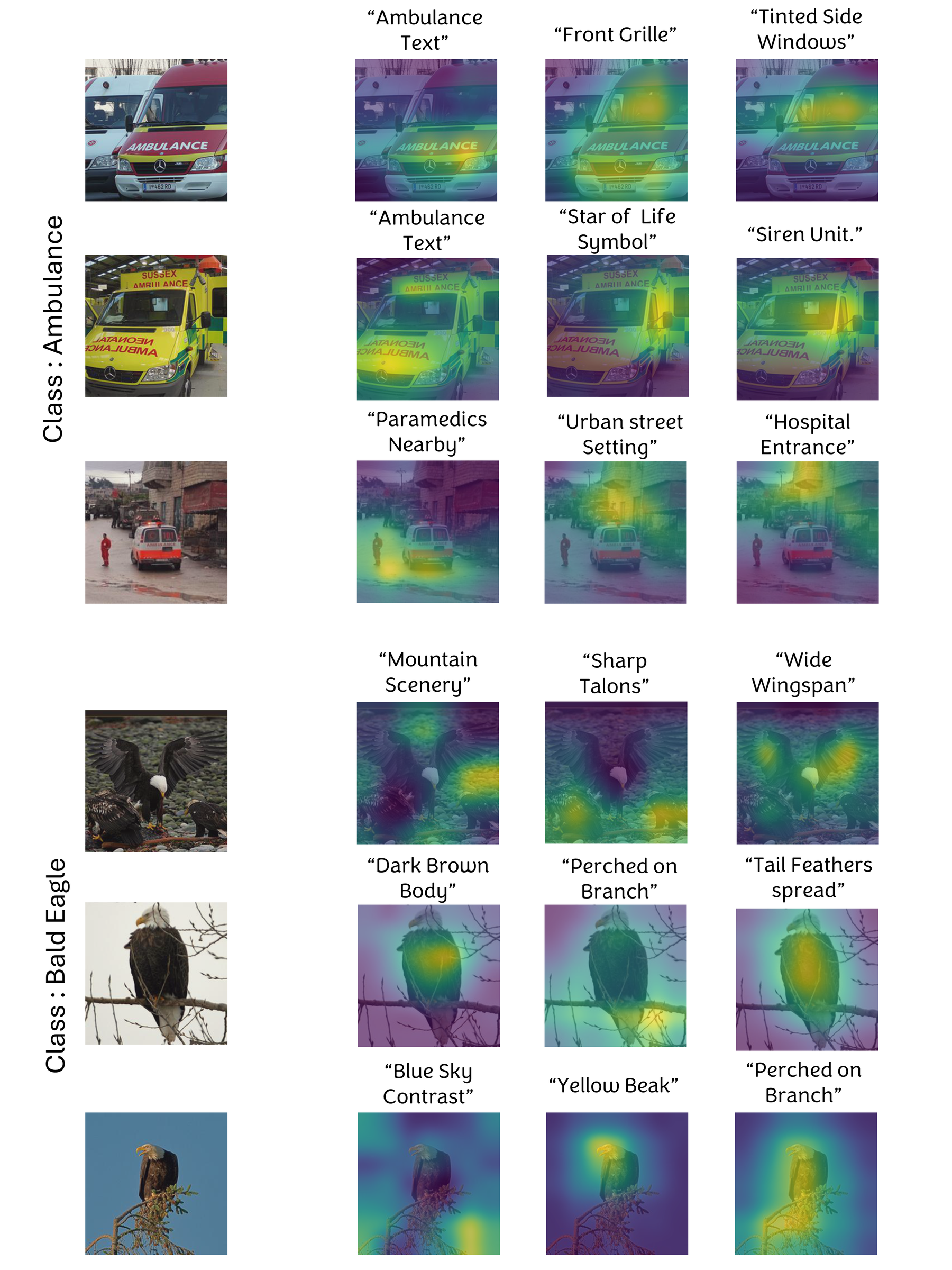}
    \caption{Concept maps for the Ambulance and Bald Eagle classes, showing input images and the top activated visual concepts.C}
    \label{fig:qualitative_ours_ambulance_bald_eagle}
\end{figure*}

\begin{figure*}[t!]
    \centering
    \includegraphics[width= .9\textwidth ]{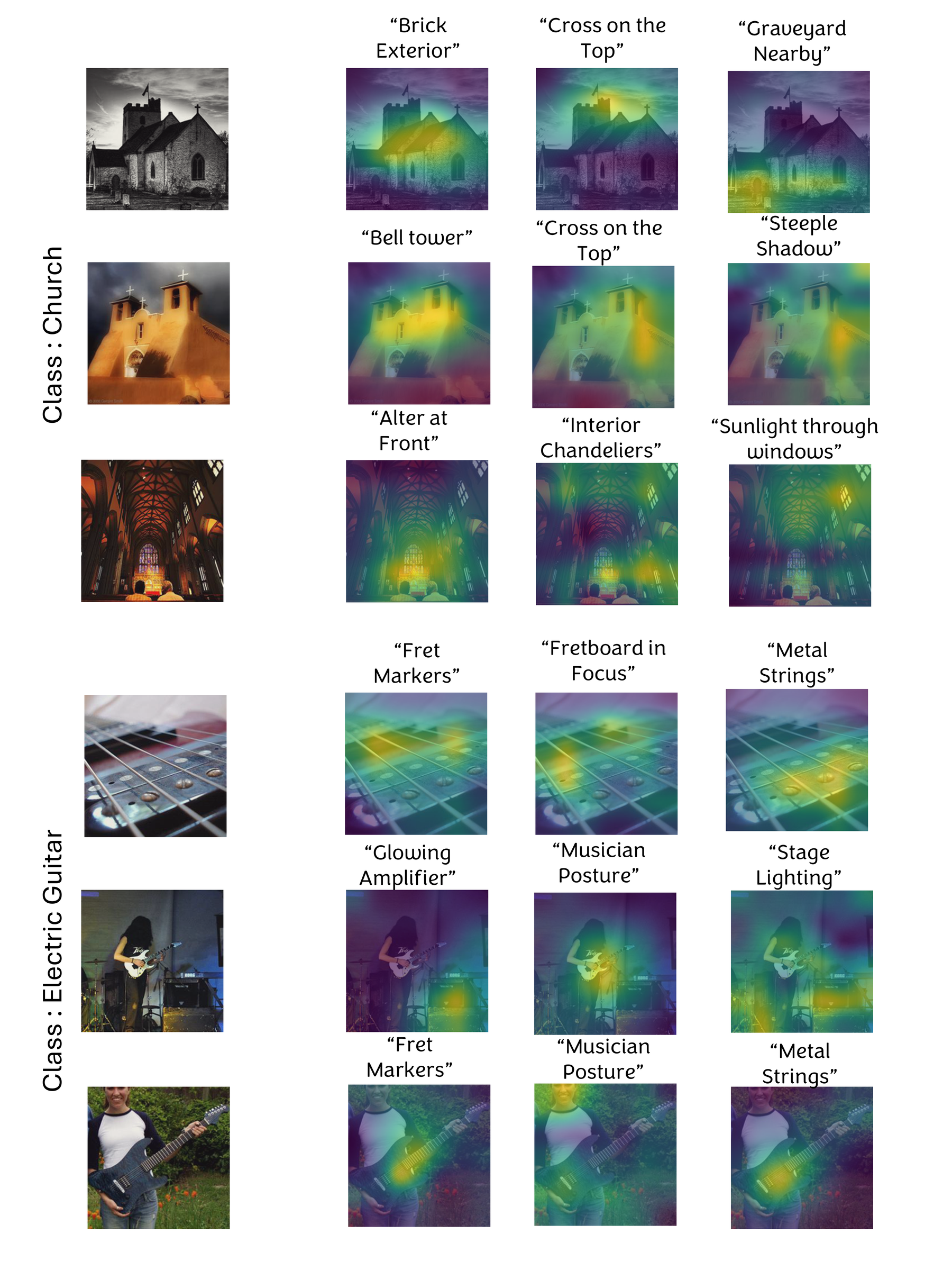}
    \caption{
Concept maps for the Church and Electric Guitar classes, highlighting the learned part-, texture-, and context-level concepts.
}

    \label{fig:qualitative_ours_chrch_guitar}
\end{figure*}

The concept visualizations in Figures~\ref{fig:qualitative_ours_banana_basketball}--\ref{fig:qualitative_ours_chrch_guitar} demonstrate that 
our method discovers interpretable and spatially grounded concepts across a diverse 
range of ImageNet categories. For object-centric classes such as \emph{Ambulance}, 
the model captures distinctive parts including “Ambulance Text,” “Front Grille,”  with heatmaps closely aligned to the 
corresponding regions of the vehicle. Contextual cues such as “Urban Street 
Setting” and “Paramedics Nearby” also emerge, reflecting the model’s ability to 
represent both object parts and scene-level signals.

For natural categories like \emph{Bald Eagle}, the discovered concepts emphasize 
fine-grained attributes such as “Sharp Talons,” “Wide Wingspan,” “Tail Feathers 
Spread,” and “Perched on Branch,” demonstrating that the factorization captures 
species-dependent morphology and pose-specific details.

The \emph{Banana} and \emph{Basketball} classes similarly exhibit semantically 
coherent concepts: color and ripeness cues (“Ripe Spot,” “Yellow Peel”), cluster 
structure (“Hanging Bunch”), and environment attributes (“Crowd in the 
Background,” “Court Floor Line,” “Player Shoes”). These maps show that 
the method cleanly separates object features from contextual scene information.

Finally, the \emph{Church} and \emph{Electric Guitar} classes highlight the 
model’s ability to encode structural and material properties. Concepts such as 
“Brick Exterior,” “Bell Tower,” and “Interior Chandeliers” reflect architectural 
components, while guitar-related concepts such as “Fret Markers,” “Metal Strings,” 
“Musician Posture,” and “Stage Lighting” localize  on instrument parts 
and performance settings.

Overall, these qualitative results illustrate that the proposed model recovers 
rich, human-understandable concepts that respect object structure, texture, and 
scene context, providing spatially faithful explanations without manual labeling.

\section{Additional Concept Discovery Comparison}\label{apendix:concept_comparison}
\begin{figure*}[t]
\vspace{-1.5mm}
    \centering
    \includegraphics[width=\textwidth]{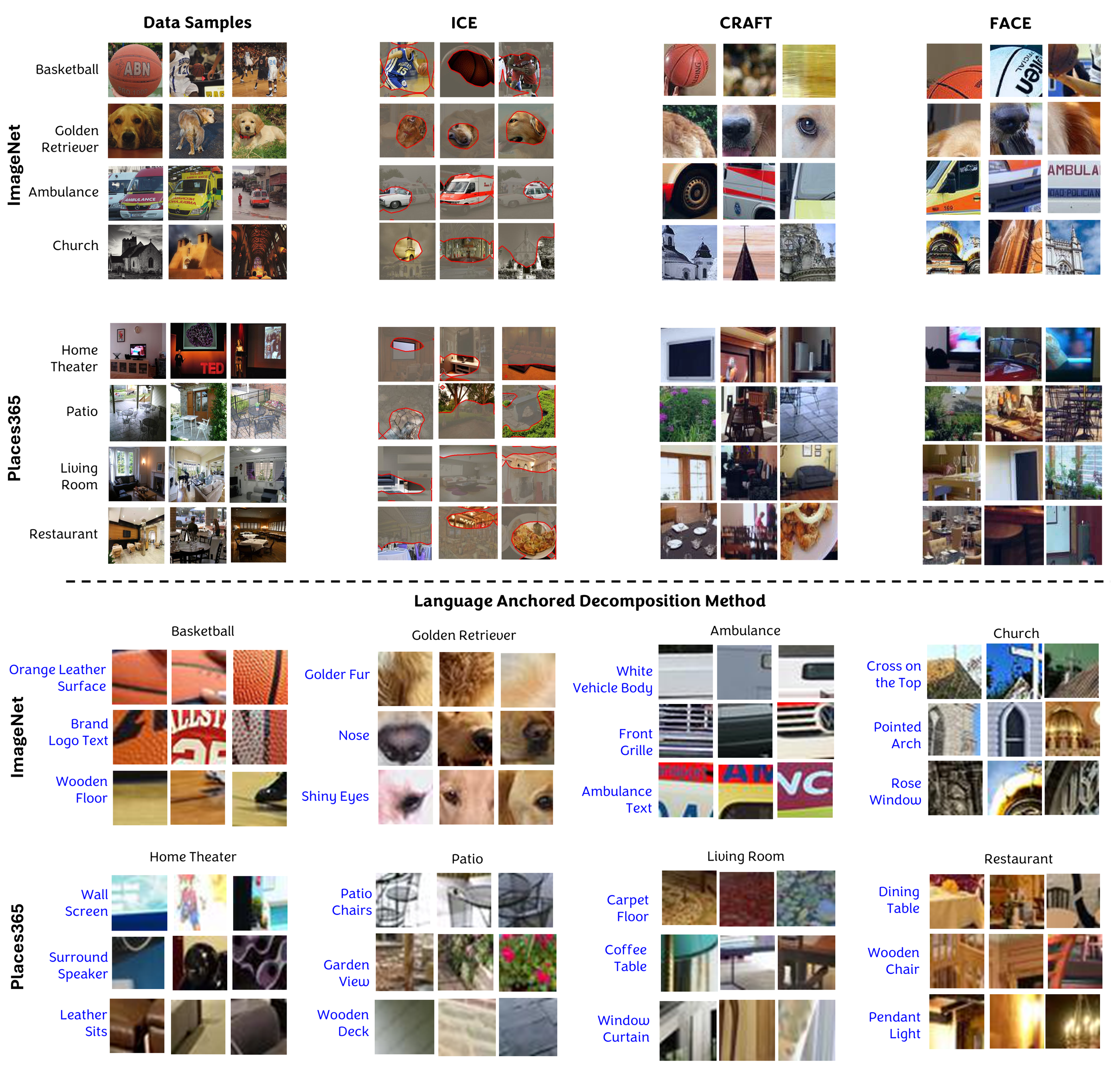}
    \caption{
Additional qualitative comparison of concept discovery methods on ImageNet and
Places365. For each class, we show representative input samples (left), followed by
concepts extracted using ICE, CRAFT, and FACE, and the concepts discovered by
\method{} (bottom). Baseline methods highlight coarse or fragmented regions, while
\method{} identifies fine-grained, semantically meaningful, and spatially coherent
concepts.}

    \label{fig:additional_comparison}
    \vspace{-3mm}
\end{figure*}

Figure~\ref{fig:additional_comparison} provides further qualitative comparisons across both 
ImageNet and Places365. ICE, CRAFT, and FACE typically focus on coarse object regions or isolated 
texture cues, often without semantic consistency across samples. In contrast,\method{} discovers interpretable, localized, and semantically named concepts—
such as part-level attributes, materials, textures, and scene elements—demonstrating clearer 
semantic grounding and improved interpretability across both object-centric and scene-centric datasets.
% {
%     \small
%     \bibliographystyle{ieeenat_fullname}
%     \bibliography{main}
% }

\end{document}